\documentclass[journal]{IEEEtran}

\IEEEoverridecommandlockouts                              
\usepackage{graphics} 
\usepackage{epsfig} 
\usepackage{times} 
\usepackage{amsmath} 
\usepackage{amssymb}  
\usepackage{bm}
\usepackage[bookmarks=true]{hyperref}
\usepackage{xcolor}
\usepackage{color}
\usepackage[ruled,vlined]{algorithm2e}
\usepackage{balance}
\usepackage{array}
\usepackage{siunitx}  
\usepackage{booktabs}
\usepackage{multirow}
\usepackage{subcaption}
\usepackage{cite}
\usepackage{mwe} 
\usepackage{soul}
\usepackage{layout}

\newcommand{\change}[1]{{\color{black}#1}}
\newcommand{\bs}{\boldsymbol}
\newcommand{\mb}{\mathbf}

\definecolor{Gray}{gray}{0.9}
\definecolor{somegray}{rgb}{0.5, 0.5, 0.5}
\newcommand{\darkgrayed}[1]{\textcolor{somegray}{#1}}

\usepackage{etoolbox}

\makeatletter
\newcommand*\titleheader[1]{\gdef\@titleheader{#1}}
\AtBeginDocument{%
  \let\st@red@title\@title
  \def\@title{%
    \vskip-2.0em
    \bgroup\normalfont\large\centering\@titleheader\par\egroup
    \vskip1em\st@red@title}
}

\makeatother 

\titleheader{ \darkgrayed{This paper has been accepted for publication in the 
\\ IEEE Transactions on Robotics (T-RO), 2021. 
\copyright IEEE} }

\title{ 
Policy Search for Model Predictive Control \\ with
Application to Agile Drone Flight
}

\author{Yunlong Song, Davide Scaramuzza
    \thanks{The authors are with the Robotics and Perception Group, Department of Informatics, University of Zurich, and Department of Neuroinformatics, University of Zurich and ETH Zurich, Switzerland (\protect\url{http://rpg.ifi.uzh.ch}).
    This work was supported by the National Centre of Competence in Research (NCCR) Robotics through the Swiss National Science Foundation (SNSF) and the European Union’s Horizon 2020 Research and Innovation Programme under grant agreement No. 871479 (AERIAL-CORE) and the European Research Council (ERC) under grant agreement No. 864042 (AGILEFLIGHT).
    }
}

\begin{document}

\maketitle

\begin{abstract}
Policy Search and Model Predictive Control~(MPC) are two different paradigms for robot control:
policy search has the strength of automatically learning complex policies using experienced data,
while MPC can offer optimal control performance using models and trajectory optimization. 
An open research question is how to leverage and combine the advantages of both approaches.
In this work, we provide an answer by using policy search for automatically choosing high-level decision variables for MPC, 
which leads to a novel \textit{policy-search-for-model-predictive-control framework}. 
\change{
Specifically, we formulate the MPC as a parameterized controller, where the hard-to-optimize
decision variables are represented as high-level policies.
Such a formulation allows optimizing policies in a self-supervised fashion. 
We validate this framework by focusing on a challenging problem in agile drone flight: 
flying a quadrotor through fast-moving gates.
Experiments show that our controller achieves robust and real-time control performance in both simulation and the real world. 
The proposed framework offers a new perspective for merging learning and control.  
}
\end{abstract}


~\\
\textbf{Code:} \url{https://uzh-rpg.github.io/high_mpc} \\
\textbf{Video:} \url{https://youtu.be/Qei7oGiEIxY}



\section{Introduction}
\label{section: intro}
\change{
Mobile robots operate in a dynamic world. 
Notably, quadrotors are agile robots that can navigate at high speeds in
highly complex and dynamic environments otherwise inaccessible to humans.}
However, sudden environmental changes, like dynamic obstacles, can raise fundamental problems 
for the vehicle control since they require the vehicle to have fast reactions and 
replan its trajectory quickly. 
An essential requirement for agile drone flight in dynamic environments is to
adapt the vehicle trajectory rapidly depending on the environmental changes.
State-of-the-art model-based approaches have shown to be effective for 
controlling the quadrotor in both static and dynamic 
environments~\cite{falanga2020dynamic, mellinger2011minimum, neunert2016fast, mueller2015computationally, Zhou19arxiv, karaman2012high, loianno2016estimation,allen2019real, richter2016polynomial, landry2016aggressive}.
For example, in the context of drone racing~\cite{foehn2020alphapilot, loquercio2019deep, moon2019challenges}, 
the drone has to fly through a sequence of static gates (subjected to small disturbance) at extremely high speeds. 
\begin{figure}[htp]
  \centering
  \includegraphics[width=0.5\textwidth]{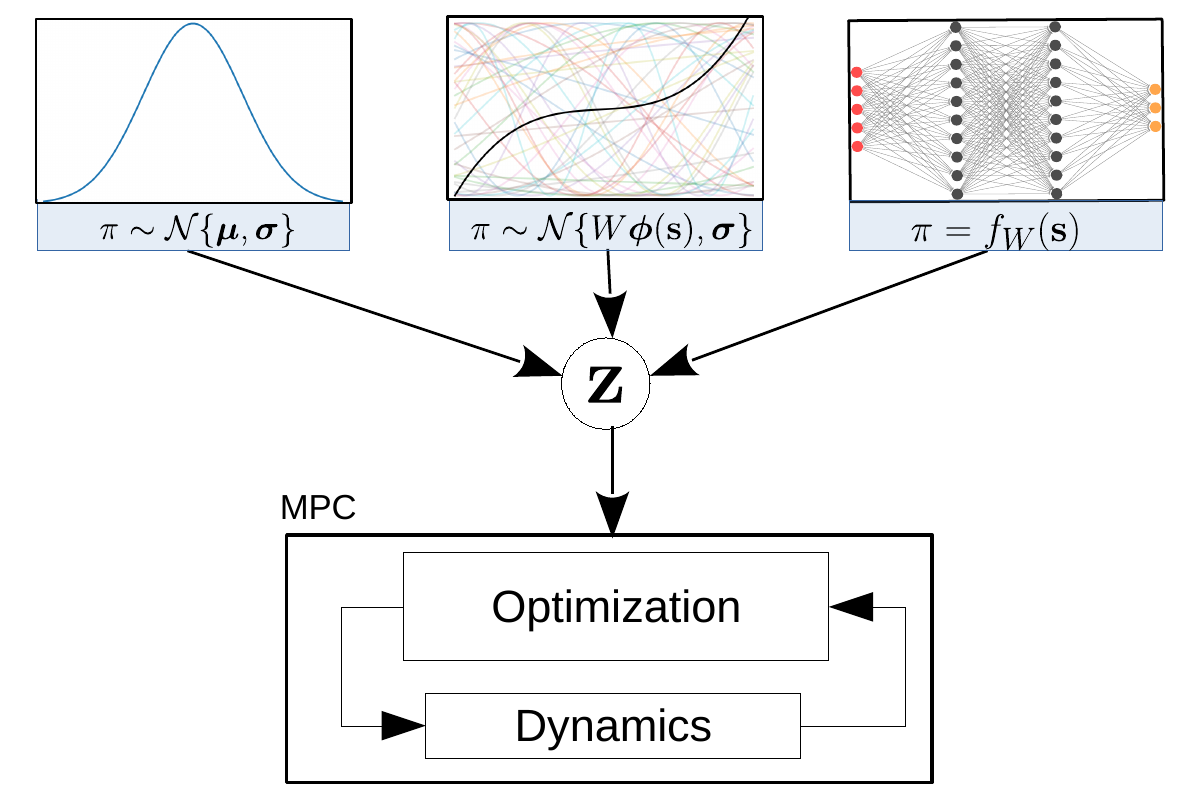}
  \caption{An overview of the proposed \textit{policy-search-for-model-predictive-control framework}. }
 \label{fig: method_overview}
\end{figure}

\begin{figure}[htp]
  \centering
  \includegraphics[width=0.5\textwidth]{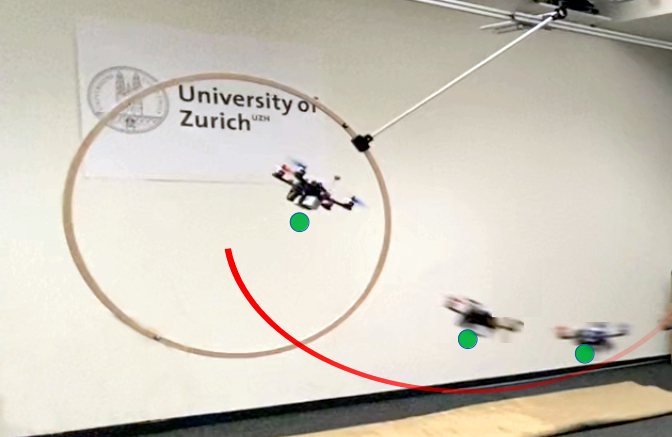}
  \caption{\change{An application of the proposed method for flying a quadrotor
  through a fast moving gate.}}
  \label{fig: fly_through}
\end{figure}

\change{
Model predictive control~(MPC)~\cite{rawlings2009model} has been shown to be a powerful 
model-based approach for solving complex quadrotor control problems~\cite{Falanga2018, neunert2016fast, kamel2017model, nguyen2020model, foehn2020cpc}.
For example, the perception-aware MPC~\cite{Falanga2018} is a framework that unifies 
both planning and perception objectives.
MPC is increasingly gaining popularity in many robotic domains,
thanks to its capability of simultaneously dealing with complex 
nonlinear dynamic systems while satisfying different state and input constraints.

Despite the successes, many MPC applications still experience significant challenges, 
such as the requirement of an accurate mathematical model and the necessity of
solving trajectory optimization problems online with the limited computational power of small-scale computers. 
In practice, the closed-loop performance of MPC for a specific task is sensitive
to several design choices, including cost function formulation, hyperparameters, and the prediction horizon.
As a result, a series of approximations, heuristics, and parameter tuning is employed, producing sub-optimal solutions.}

\change{
On the other hand, reinforcement learning~(RL)~\cite{sutton2018reinforcement} methods, 
like policy search, allow solving
continuous control problems with minimum prior knowledge about the task. 
The key idea of RL is to automatically train the policy via trial and error and maximize the task performance measured by the given reward function. 
While RL has achieved impressive results in solving a wide range of robot control tasks~\cite{hwangbo2017control, lee2020learning, song2021overtaking, song2021autonomous}, the lack of interpretability of an end-to-end controller trained using RL is of significant concern by the control community~\cite{donti2020enforcing}.}

Ideally, the control framework should be able to combine the advantages of both methods---the ability of model-based controllers, like MPC,
to safely control a physical robot using the well-established knowledge in dynamic modeling and optimization and the power of RL to learn complex policies using experienced data automatically.
Therefore, the resulting control framework can handle large-scale inputs, reduce human-in-the-loop design and tuning, and eventually achieve adaptive and optimal control performance. 
Despite these valuable features, designing such a system remains a significant challenge. 

To this end, one line of research in the learning community has been focusing on
developing data-efficient policy search methods using model priors.
For instance, guided policy search~(GPS) algorithms~\cite{zhang2016learning, levine2016end, kaufmann2020deep}
opt for transforming RL into a supervised learning problem.
The key idea in GPS is to use a trajectory optimization algorithm to collect training data
for training neural networks via supervised learning.
However, these methods still learn black-box control policies that suffer from poor generalizations. 
The second trend pertains to learning-based MPC~\cite{williams2017infoMPC, kabzan2019learning, ostafew2016robust, rosolia2019learning}, 
which can leverage real-world data to improve dynamics modeling and use model predictive path integral control~(MPPI)~\cite{williams2017model} for optimization.
Such algorithms generally have their roots in stochastic optimal control and require sampling a large amount of data in real-time for optimization, making those methods computationally expensive. 

\subsection*{Contributions}
\change{
In this work, we propose a new paradigm for merging learning and control: 
learning high-level policies for model predictive control using policy search. 
An overview of our approach is summarized in~Fig.~\ref{fig: method_overview}.
Specifically, we consider the MPC as a parameterized controller
and formulate the search of high-level decision variables for MPC as a probabilistic policy search problem.
First, we use two general Gaussian policies for modeling the high-level decision variables 
and show that the policy updates have closed-form solutions. 
Second, we propose a self-supervised training method for learning neural network policies. 
Our key insight is that policy search is useful for making
high-level decisions for MPC, allowing automatically learning 
and adapting hard-to-optimize parameters.
}

\change{On the experiment side, we evaluate our approach 
by addressing a challenging problem towards autonomous agile drone flight in dynamic environments:
controlling a quadrotor to fly through a sequence of fast-moving gates.
The key advantage of our approach compared to the standard MPC formulation is that the desired traversal time,
which is hard to optimize simultaneously with other state variables, can be learned offline and can be adaptively
selected at runtime.
The resulting controller, which consists of a trained neural network policy and an MPC,
achieves real-time control performance of a physical drone.
An illustration of the real-world experiment is shown in~Fig.~\ref{fig: fly_through}.
}

This work is an extension of our previous conference paper~\cite{Yunlong2020learning}.
The earlier version of this work proposed learning Gaussian and neural network policies and demonstrated learning a single time variable for flying a quadrotor through a dynamic gate in simulation. 
In this paper, we additionally   
1) introduce a new algorithm for learning a Gaussian linear policy,
2) demonstrate that our approach is a general framework that can  
learn multidimensional decision variables, not just a single variable, 
3) demonstrate that our controller can control a drone to fly through multiple gates in simulation and outperforms a standard MPC and a trajectory sampling method,
4) deploy the algorithm on a physical drone and show that the trained neural network high-level policy can be transferred to the real world without fine-tuning.

\section{Related Work}
\subsection{Policy Search for Robotics}
Policy search~\cite{sutton2018reinforcement} is a central area of reinforcement learning concerned with
how to find an optimal parametric policy by maximizing the expected 
return of sampled trajectories. 
Depending on their exploration strategies for the stochastic trajectory generation,
policy search methods can be categorized into step-based and episode-based methods~\cite{deisenroth2013survey, sutton2018reinforcement}.
Most variations of policy search methods make use of step-based exploration strategies 
by adding different exploration noise in the \textit{action space} at each control time step.
Step-based policy search algorithms~\cite{schulman2015trpo, williams1992simple, kakade2001natural, peters2010reps} are widely used for continuous control tasks, ranging from learning agile 
motor skills for legged robots~\cite{hwangbo2019learning} to controlling a simulated race car 
at its friction limits~\cite{song2021overtaking}.
They learn end-to-end black-box control policies that can map observations 
directly to control commands. 

By contrast, episode-based policy search methods~\cite{deisenroth2013survey, stulp2012path, sun2009efficient, sehnke2008policy} 
add exploration noise in the \textit{parameter space} of the policy only at the beginning of the episode. 
In particular, episode-based methods are widely used for learning movement primitives~\cite{schaal2006dynamic, paraschos2013probabilistic, williams2008modelling}, which are compact parameterizations of the robot's control policy.
For example, the task-parameterized dynamic motor primitives~(DMPs)~\cite{schaal2006dynamic, ijspeert2013dynamical} are popular 
compact policy representations in robotics.
Adjusting their parameters allows robots to learn new skills quickly and solve many challenging robot control problems,
such as playing \textit{Baseball}~\cite{peters2008reinforcement}, \textit{Ball-in-the-cup}~\cite{kober2009policy}, and \textit{Table Tennis}~\cite{kober2011adjust}. 
Episode-based policy search methods help learn compact skills representations that are not easy to model by human experts.

\subsection{Data-driven Control with Model Predictive Control}
\subsubsection{MPC-guided Policy Search}
Model-free policy search algorithms learn control policies via trial-and-error; however, they suffer from high sample complexities. 
Guided policy search~\cite{levine2013guided} converts model-free policy search to supervised learning by iteratively collecting the training data using offline trajectory optimization~\cite{levine2014learning1, levine2016end, levine2013guided, levine2014learning2} or 
model predictive control~\cite{zhang2016learning, kaufmann2020deep}. 
A key advantage of guided policy search is that it effectively trains deep neural network control policies, where the policy can handle complex and high-dimensional inputs from onboard sensors. 
For example, a deep sensorimotor policy, trained using MPC and imitation learning, enables an autonomous quadrotor to fly extreme acrobatic maneuvers with only onboard sensing and computation~\cite{kaufmann2020deep}.
However, this line of work usually only uses the model during training and results in a policy specialized in a single task. 
Despite all of the successes achieved by guided policy search, 
the lack of generalization and robustness of the end-to-end policy remains a primary challenge.

\subsubsection{Learning-based MPC}
In the second paradigm, learning-based MPC~\cite{kabzan2019learning, williams2018information, williams2017infoMPC, ostafew2016robust, rosolia2019learning} 
can leverage real-world data to improve dynamic modeling or learn a cost function for MPC. 
It allows for a more robust and flexible MPC design.
In particular, sampling-based MPC~\cite{williams2018information} algorithms
are developed for handling complex cost criteria and general nonlinear dynamics.
This is achieved by combining neural networks for the system dynamics approximation with
the model predictive path integral (MPPI) control framework~\cite{williams2018information} for real-time control optimization. 
A crucial requirement for the sampling-based MPC is to generate a large number of samples 
in real-time, where the sampling procedure is generally performed in parallel 
by using graphics processing units~(GPUs).
Hence, it is computationally and memory expensive to run sampling-based MPC on embedded systems.
These methods generally focus on learning dynamics for tasks where a dynamical model of the robots
or its environment is challenging to derive analytically, such as 
aggressive autonomous driving around a dirt track~\cite{williams2017infoMPC}.

Alternatively, differentiable MPC~\cite{amos2018differentiable} treats the MPC as a 
differentiable policy class for reinforcement learning. 
Hence, by differentiating through the optimization problem using the Karush–Kuhn–Tucker~(KKT) 
conditions of the convex approximation at a fixed point of the controller, 
it can also learn the costs and dynamics of an MPC controller via end-to-end learning. 
The analytical derivative relies on a fixed point of the controller, which, however, often does not exist 
when using neural networks to approximate the dynamics~\cite{amos2018differentiable}.

\begin{figure}[t!]
     \centering
     \includegraphics[width=0.5\textwidth]{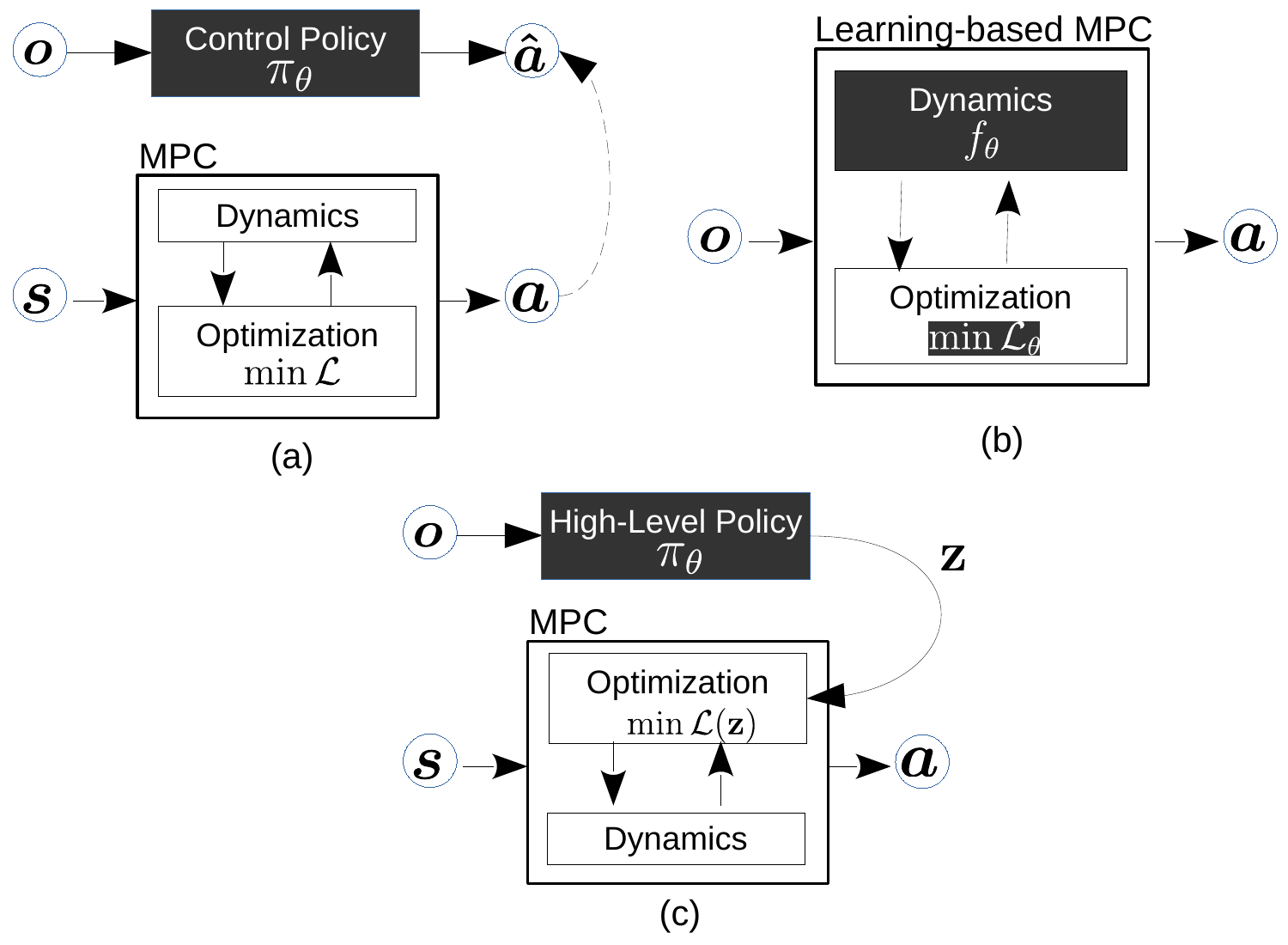}
     \caption{Taxonomy of existing methods combining machine learning with model predictive control.}
     \label{fig: learningmpc}
\end{figure}

\subsubsection{Learning Neural Network Policies for MPC}
To combine the power of neural networks and the strength of standard MPC optimization, state-of-the-art
systems~\cite{kaufmann2019beauty, drews2017aggressive, drews2019vision} opt for using
deep neural networks as standalone representation learning modules.
Specifically, a neural network is trained to process high-dimensional data, such as images, and is used to predict low-dimensional state information for the MPC.
For instance, \cite{kaufmann2019beauty} proposed to combine a Convolutional Neural Network~(CNN) 
with an MPC controller to solve the problem of navigating a quadrotor to pass through multiple gates.
The trained neural network predicts three-dimensional poses of the gate's center from image observations, and then, 
the MPC outputs control commands for the quadrotor to navigate through the predicted waypoints.
Similarly, the method in \cite{drews2017aggressive} tackles an aggressive autonomous driving problem by using a CNN-based policy to predict a cost map of the track, which is then directly used for online trajectory optimization. 
A key advantage of this line of work is that it combines the benefits of both neural networks for high-dimensional data processing and MPC for robot control.

%

\section{Preliminary}
\label{sec: preliminary}
We introduce mathematical formulations for both MPC and policy search.
We discuss three kinds of policy representations that are widely used in RL
and use them to model the high-level policies.

\subsection{Model Predictive Control}
We consider the problem of controlling a nonlinear deterministic dynamical system 
whose dynamics is defined by a differential equation $\mb{\dot{x}}_t = f(\mb{x}_t, \mb{u}_t)$,
where $\mb{x}_t \in \mathbb{R}^{n}$ is the state vector, $\mb{u}_t \in \mathbb{R}^{m}$ is 
a vector of control commands, and $\mb{\dot{x}}_t \in \mathbb{R}^{n}$ is the derivative of the current state.
In MPC, we approximate the actual continuous time differential
equation using a set of discrete time integration $\mb{x}_{h+1}= \mb{x}_h + d_t \cdot \hat{f}(\mb{x}_h, \mb{u}_h)$,
with $d_t$ as the time interval between consecutive 
states and $\hat{f}$ as an approximated dynamical model. 

Let $\mb{r} \in \mathbb{R}^{k}$ be a vector of reference states, e.g., a planned trajectory. 
We define a vector of high-level decision variables as $\mb{z} \in \mathbb{R}^{N}$.  
For example, for our application (Section~\ref{sec: application_preliminary}.),
$\mb{z}=[z_1, \cdots, z_N]$ defines the time variables at which the robot should be passing 
the corresponding gate. 
At every control time step $t$, the system is in state $\mb{x}_t$. 
MPC takes the current state $\mb{x}_\text{init} = \mb{x}_t$, the reference states $\mb{r}$,
and the high-level decision vector $\mb{z}$ as input. 

Formally, MPC minimizes a cost function over a fixed finite time horizon $H$ by solving an optimization problem:
\begin{equation}
    \label{eq: mpc_obj}
    \begin{aligned}
     &\min_{ \mb{u}_{1:H}, \mb{x}_{1:H} } & \mathcal{L} = \sum_{h=1}^{H} c(\mb{x}_h, \mb{u}_h; \mb{r}, \mb{z}) \\
     &\text{subject to} & \quad \mb{g}(\mb{x}, \mb{u}) = 0, \quad \mb{h}(\mb{x}, \mb{u}) \leq 0  \\
     && \mb{x}_{h+1} = \mb{x}_h + d_t \cdot \hat{f}(\mb{x}_h, \mb{u}_h), & \quad
     \mb{x}_1 = \mb{x}_\text{init}
    \end{aligned}
\end{equation}
where $\mb{g}(\mb{x}, \mb{u})$ represents equality constraints and
$\mb{h}(\mb{x}, \mb{u})$ represents inequality constraints.
%

Our goal is to find the optimal control command $\mb{u}^{\ast}$ for the current state such
that we can execute it and move the robot to the next state.
It is achieved by solving the trajectory optimization problem in real-time
and by repeating this process at every control time step. 
Specifically, MPC minimizes the cost in the future states and 
outputs an optimal trajectory~($\bs{\tau}$) that consists of a 
sequence of control commands and states.
Only the first command $\mb{u}^{\ast}=\mb{u}_1$ is executed on the robot. 

In this work, we take the MPC as a parametric controller that is parameterized by the high-level
decision variables $\mb{z}$. 
Therefore, modulating the variables $\mb{z}$ can result in different MPC outputs, denoted as $\bs{\tau} = \text{MPC}(\mb{z})$.
For example, in the context of flying through a dynamic gate~(Section~\ref{sec: application_preliminary}.), 
$\mb{z}$ can be the desired time at which the vehicle passes through the gate. 
This formulation allows us to incorporate reinforcement learning as well as different 
function representations into the MPC design. 

\subsection{Policy Search}
\change{We summarize policy search by following the derivation from~\cite{deisenroth2013survey}, in particular, 
we focus our discussion on episode-based policy search (or episodic reinforcement learning).}
Unlike many step-based policy search algorithms, which explore the action space by 
adding exploration noise directly to the policy output, 
episode-based policy search adds perturbations in the policy parameter space.
This kind of exploration is normally added at the beginning of an episode and a reward function~$R(\bm{\tau})$
is used to evaluate the quality of trajectories~$\bm{\tau}$ that are generated by sampled parameters~$\bm{\theta}$.
A comprehensive survey and tutorial about different policy search algorithms can be found here~\cite{chatzilygeroudis2019survey, deisenroth2013survey}.

Policy search algorithms try to update the policy parameters~$\bm{\theta}$ by maximizing the 
expected return of sampled trajectories
\begin{equation}
    J_{\bm{\theta}} =\mathbb{E}[R(\bm{\tau}) | \bm{\theta}] \approx \int R(\bm{\tau}) p_{\bm{\theta}}(\bm{\tau}) d\bm{\tau}.
\end{equation}
A list of episode-based policy search algorithms have been discussed in the
literature, such as policy gradient methods~\cite{kohl2004policy, williams1992simple, peters2006policy}, 
expectation-maximization~(EM) methods~\cite{kober2009policy}, 
and information-theoretic methods~\cite{christian2016hreps, peters2010reps}.

Policy gradient methods use gradient-ascent for maximizing the expected return and are simple to implement. 
However, it requires manual selection of learning rates and has an unstable learning process or slow convergence. 
Information-theoretic approaches rely on solving constrained optimization for maximizing the objective, and at the same time, they constrain the information loss by bounding the Kullback-Leibler~(KL) divergence between the new policy and the old policy. 
The requirement of solving constrained optimization limits the usage of information-theoretic algorithms to solve high-dimensional problems, such as optimizing neural network policies, making it challenging to implement in reality. 

On the other hand, the EM-based policy search algorithms provide closed-form solutions for many commonly used policy representations, and hence, do not require the user to specify the learning rate.
In addition, they provide a good trade-off between computational efficiency and sample complexity. 
This is realized by formulating policy search as a probabilistic inference problem with latent variables, which leads to a weighted maximum likelihood estimate. 
Subsequently, we can use the Expectation-Maximization algorithm to update the policy parameters. 
We focus on a probabilistic model in which the search for high-level decision variables in the MPC optimization is treated as a probabilistic inference problem. 

\subsection{Policy Representations}
We represent the high-level policy as~$\pi_{\bm{\theta}}$, which is modeled as a probability distribution 
or a deterministic policy (e.g., a neural network),
and use the policy to select high-level decision variables~$\mathbf{z} \sim \pi_{\boldsymbol{\theta}}$. 
Here, $\bm{\theta}$ are the policy parameters that have to be trained. 

\subsubsection{Gaussian Policy, Constant Mean} \label{sec: pi_gp}
First, we consider a simple scenario where the goal is to find a set of fixed decision variables~$\mathbf{z}$.
The variables are independent of the robot's state.
We use a Gaussian distribution~$\mathbf{z} \sim \pi_{\bm{\theta}}=\mathcal{N}(\bm{\mu}, \bm{\Sigma})$
to represent the policy, where $\bm{\mu}$ is a mean vector and 
$\bm{\Sigma}$ is a diagonal covariance matrix. 
The covariance matrix is needed in order to incorporate exploration. 
Therefore, the policy parameters are~$\bm{\theta}=[\bm{\mu, \Sigma}]$. 

\subsubsection{Gaussian Policy, Linear Mean} \label{sec: pi_gplinear}
Second, we consider a more general problem in which we want to find a set of adaptive decision variables,
denoted as~$\mathbf{z} = f(\mathbf{s})$.
The decisions variables are associated with the robot's context~$\mathbf{s}$.
We use a Linear Gaussian model~$\mathbf{z} \sim \pi_{\bm{\theta}}=\mathcal{N}(\mathbf{W}\phi(\mathbf{s}), \bm{\Sigma})$
to denote the policy, in which the Gaussian mean $\bm{\mu} = \mathbf{W}\phi(\mathbf{s})$ is 
represented by a linear function approximator, linear with respect to the function parameters~$\mathbf{W}$.
Here, $\phi: \mathbf{s} \subset \mathbb{R}^{N} \rightarrow  \mathbb{R}^{M}$ is a kernel featurizer 
that converts the states of dimension~$N$ into a vector of features of dimension~$M$ using
basis functions, such as Radial Basis Functions~(RBF)~\cite{sutton2018reinforcement} or 
Random Fourier Features~(RFF)~\cite{rajeswaran2017towards}.
Therefore, the policy parameters are ~$\bm{\theta} = [\mathbf{W}, \bm{\Sigma}]$. 

\subsubsection{Neural Network Policy} 
\label{sec: pi_nn}
We use a neural network $\bm{z}_t = f_{\boldsymbol{\theta}}(\mathbf{o}_t)$ as a deterministic policy representation.
Here, $f_{\boldsymbol{\theta}}$ represents the neural network and $\mathbf{o}_t$ is the robot's observation at different time step~$t$. 
The solution for updating the parameters~$\boldsymbol{\theta}$ of a neural network in policy search is difficult to derive analytically due to the highly nonlinear property of neural networks. 
Many deep reinforcement learning algorithms are based on policy gradients~\cite{henderson2018deep},
which are known to have unstable learning processes or slow convergence. 
By contrast, we use a self-supervised learning algorithm for training the neural network policy (Algorithm~\ref{algo: neuralmpc}).

\section{Probabilistic Policy Search for MPC}
\label{sec: ps4mpc}

\subsection{Problem Formulation}
\begin{figure}[!htp]
     \centering
     \includegraphics[width=0.3\textwidth]{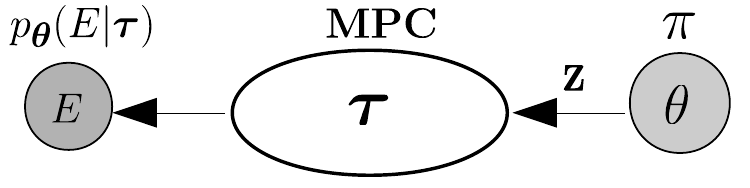}
         \caption{A graphical model of probabilistic policy search for model predictive control.}
     \label{fig: vi}
\end{figure}

We treat MPC as a controller~$\bm{\tau} =\text{MPC}(\mathbf{z})$ that is
parameterized by the high-level decision variables~$\bm{z}$. 
Here, $\bm{\tau} = [\mathbf{u}_h, \mathbf{x}_h]_{h\in{1, \cdots, H}}$ is a trajectory generated by MPC given~$\mathbf{z}$, where
$\mathbf{u}_h$ are control commands and $\mathbf{x}_h$ are corresponding states of the robot. 
By perturbing~$\mathbf{z}$, MPC can result in completely different trajectories $\bm{\tau}$.
To find the optimal trajectory for a given task, the optimal $\mathbf{z}$ has to be defined in advance.
First, we model~$\mathbf{z}$ as a high-level policy represented by a probability distribution, 
specifically a parameterized Gaussian distribution. 
Then, we optimize the policy using probabilistic policy search (or probabilistic inference) algorithms. 
A visualization of the inference problem is given in Fig~\ref{fig: vi} (inspired by~\cite{deisenroth2013survey}).

To formulate the policy search as a latent variable inference problem,
similar to~\cite{neumann2011variational, toussaint2009robot, deisenroth2013survey},
we introduce a binary ``reward event” as an observed variable, denoted as $E = 1$.
Maximizing the reward signal implies maximizing the probability of this ``reward event”.
The probability of this reward event is given by~$p(E | \bm{\tau}) \propto \exp{ \{ R(\bm{\tau}) \} }$,
where $R(\bm{\tau})$ is a reward function for evaluating
the goodness of the MPC solution $\bs{\tau}$ with respect to a given evaluation metric of the task.
This leads to the following maximum likelihood problem~\cite{deisenroth2013survey}:
\begin{equation}\label{eq: max_log_pro}
    \max_{\bs{\theta}} \quad \log p_{\bs{\theta}}(E=1) = \log \int_{\bs{\tau}} p(E|\bs{\tau}) p_{\bs{\theta}} (\bs{\tau)} d \bs{\tau},
\end{equation}
which is intractable to solve directly and can be approximated efficiently using 
Monte-Carlo Expectation-Maximization~(MC-EM)~\cite{kober2009policy, vlassis2009model}. 
MC-EM algorithms find the maximum likelihood solution for
the log marginal-likelihood~(\ref{eq: max_log_pro}) by introducing a variational distribution~$q(\bs{\tau})$,
and then, decompose the marginal log-likelihood into two terms:
\begin{equation}
    \log p_{\bs{\theta}}(E=1) = \mathcal{L}_{\bs{\theta}} (q(\bs{\tau}) ) + D_\text{KL}(q(\bs{\tau}) || p_{\bs{\theta}}(\bs{\tau} | E) )
\end{equation}
where the~$D_\text{KL}$ is the Kullback–Leibler~(KL) divergence 
between $q(\bm{\tau})$ and the reward-weighted trajectory distribution~$p_{\bs{\theta}}(\bs{\tau} | E)$. 
Here, $\mathcal{L}_{\bs{\theta}} (q(\bs{\tau}))$ is the lower 
bound of~$\log p_{\bs{\theta}}(E=1)$ as $D_\text{KL} \geq 0$.

The MC-EM algorithm is an iterative method that alternates between 
performing an Expectation~(E) step and a Maximization~(M) step.
In the expectation step, we minimize the KL-divergence~$D_\text{KL}$,
which is equivalent to setting $q(\bs{\tau})=p_{\bs{\theta}}(\bs{\tau} | E) \propto p(E|\bs{\tau})p_{\bs{\theta}}(\bs{\tau})$.
In the maximization step, we update the policy parameters by maximizing the 
expected complete data log-likelihood
\begin{equation}
    \bs{\theta}^{\ast} = \arg \max_{\bs{\theta}} \sum_i p(E | \bs{\tau}^{[i]}) \log p_{\bs{\theta}} (\bs{\tau}^{[i]})
\end{equation}
where each sample $\bs{\tau}^{[i]}$ is weighted by the probability of the ``reward event", denoted as $p(E|\bs{\tau})$.
The trajectory distribution $p_{\bs{\theta}} (\bs{\tau}^{[i]})$ can be replayed by 
the high-level policy~$\pi_{\bm{\theta}}$.
To transform the reward signal $R(\bs{\tau}^{[i]})$ of a sampled trajectory $\bs{\tau}^{[i]}$
into a probability distribution of the ``reward event", we use the exponential 
transformation~\cite{neumann2011variational, toussaint2009robot, deisenroth2013survey}:
\begin{equation}
    \label{eq: d}
    d^{[i]} =p(E|\bs{\tau}) = \exp{ \left\{ \beta R(\bs{\tau}^{[i]}) \right\} }
\end{equation}
where the parameter $\beta \in \mathbb{R}_{+}$ denotes the inverse temperature of the soft-max distribution,
higher value of $\beta$ implies a more greedy policy update.

\subsection{Learning Gaussian Policies}
\subsubsection{Gaussian Policy, Constant Mean}
We first focus on solving a simple problem of learning a Gaussian 
policy~$\pi_{\bm{\theta}}(\bm{z}|\bm{\mu}, \bm{\Sigma})$ whose mean is a vector of unknown variables.
We consider the robot at a fixed state $\mb{x}_0$, which does not change during learning.
At the beginning of each training iteration, we randomly sample a list of parameters of length $N$ 
from the current policy distribution~$\pi_{\bs{\theta}}$ and evaluate the parameters via a predefined
reward function $R(\bs{\tau})$, where $\bs{\tau}^{[i]}$ are the trajectories predicted by solving the MPC
given the sampled variables $\bs{z}^{[i]}$.

In the Expectation step, we transform the computed reward signal $R(\bs{\tau})$ into
a non-negative weight $d^{[i]}$~(improper probability distribution) via the exponential transformation~(\ref{eq: d}). 
In the Maximization step, we update the policy parameters by optimizing the 
weighted maximum likelihood objective:
\begin{equation}
    \label{eq: wml}
    \bs{\theta}^{\ast} = \arg \max_{\bs{\theta}} \left\{ \sum_i d^{[i]} \log \pi_{\bs{\theta}} (\bs{z}^{[i]} ) \right\}
\end{equation}
where the policy parameters, both the mean and the covariance, are updated using 
the following closed-form expressions:
\begin{equation}
    \label{eq: update_wml}
    \begin{aligned}
    \bs{\mu} &=\left(\sum_{i=1}^N d^{[i]} \mb{z}^{[i]}\right)/\left(\sum_{i=1}^N d^{[i]}\right) \\
    \bs{\Sigma} &=\left( \sum_{i=1}^N d^{[i]}(\mb{z}^{[i]}-\bs{\mu})(\mb{z}^{[i]}-\bs{\mu})^T\right)/Y, \\
    \text{where} & \\
    Y &= \left( \left(\sum_{i=1}^{N} d^{[i]} \right)^2 - \sum_{i=1}^N (d^{[i]})^2 \right) / \left(\sum_{i=1}^N d^{[i]} \right).
    \end{aligned}
\end{equation}

We repeat this process until the expectation of the sampled reward converges. 
After training (during policy evaluation), we simply take the mean vector of the Gaussian policy 
as the optimal decision variables for the MPC.
Therefore, $\mb{z}=\bs{\mu}^{\ast}$ is the optimal MPC decision variables found by our policy search.
A complete episode-based policy search for learning a high-level Gaussian policy for MPC 
is given in Algorithm~\ref{algo: gaussianmpc}.
\change{A detailed derivation of the above solution is available in the Appendix.}

\begin{algorithm}[!htp]
    \caption{{\bf Learning Gaussian Policies for MPC} 
    \label{algo: gaussianmpc}}
	\KwIn{ $\pi_{\bs{\theta}}(\bs{\mu}, \bs{\Sigma}), N, \text{MPC}, \mb{x}_0, \mb{p}$}
	\textbf{While not converged} \\
	\quad Sample variables: $\bs{z}^{[i]} \sim \pi_{\bs{\theta}}(\bs{\mu}, \bs{\Sigma})_{i=1...N}$\\
	\quad Sample trajectories: $\bs{\tau}^{[i]} = \text{MPC.solve}(\mb{x}_0, \bs{z}^{[i]}, \mb{p})$ \\
	\quad \textbf{Expectation:} \\
	\quad \quad $d^{[i]} = \exp{ \left\{ \beta R(\bs{\tau}^{[i]}) \right\}}$ \\
	\quad \textbf{Maximization:} \\
	\quad \quad $\bs{\theta}^{\ast} = \arg \max_{\bs{\theta}} \left\{ \sum_i d^{[i]} \log \pi_{\bs{\theta}} (\bs{z}^{[i]}) \right\}$ \\
	\KwOut{Trained Policy $\pi_{\bs{\theta}^{\ast}}(\bs{\mu}^{\ast}, \bs{\Sigma}^{\ast})$}
\end{algorithm}

\subsubsection{Gaussian Policy, Linear Mean}
Algorithm~\ref{algo: gaussianmpc} can learn a Gaussian policy that only
suits for a single experiment setting or a specific scenario. 
For generalizing the learned policy to different scenarios, we extend the algorithm by
learning a Gaussian linear policy~$\pi_{\bm{\theta}}(\bm{z} | \bm{s}) \sim \mathcal{N}(\mathbf{W}\bm{\phi}(\mathbf{s}), \bm{\Sigma})$ whose mean is a linear function approximator~$\bm{\mu} = \mathbf{W}\bm{\phi}(\mathbf{s})$.
\change{We characterize a scenario by the robot's context, denoted by a vector~$\mathbf{s}$.
The problem of learning~$\pi_{\bm{\theta}}(\mathbf{z} | \mathbf{s})$ is called contextual policy search and
can be defined by maximizing the expected returns over all different contexts:
}

\begin{equation}
 \label{eq: obj_linear_gaussian}
 \max_{\bm{\theta}} \int_{\bm{s}} \rho( \bm{s}) \int_{\bm{z}} \pi_{\bm{\theta}}(\bm{z} | \bm{s}) \int_{\bm{\tau}} p(\bm{\tau} | \bm{z}, \bm{o}) R(\bm{\tau}, \bm{s})
d\bm{\tau}d\bm{z}d\bm{s}
\end{equation}
where $\rho(\bm{s})$ is the distribution over~$\bm{s}$. 
The objective~(\ref{eq: obj_linear_gaussian}) can be optimized using 
the standard MC-EM algorithm, and it results in a different weighted 
maximum likelihood objective:
\begin{equation}
    \label{eq: linear_wml}
    \bs{\theta}^{\ast} = \arg \max_{\bs{\theta}} \left\{ \sum_i d^{[i]} \log \pi_{\bs{\theta}} (\bs{z}^{[i]} | \bm{s}^{[i]} \ ) \right\}.
\end{equation}

Maximizing Eq.~(\ref{eq: linear_wml}) results in closed-form solutions
for the policy parameters:
\begin{equation}
    \label{eq: sol_lineargp}
    \begin{aligned}
    \bm{W} &=\left(\bm{\Phi}^{T} \bm{D} \bm{\Phi}+\lambda \bm{I}\right)^{-1} \bm{\Phi}^{T} \bm{D} \bm{\Theta} \\
    \bm{\Sigma} &=\frac{\sum_{i=1}^{N} d^{[i]}\left(\bm{u}^{[i]}-\bm{W}^{T} \bm{\phi}\left(\bm{s}^{[i]}\right)\right)\left(\bm{u}^{[i]}-\bm{W}^{T} \bm{\phi}\left(\bm{s}^{[i]}\right)\right)^{T}}{Y},
    \end{aligned}
\end{equation}
where $\bm{\Phi}=[\phi(\bm{s})^{[i]}, \cdots, \phi(\bm{s})^{[N]}]$ is a matrix that 
contains converted feature vectors for all sampled contexts~$\bm{s}$ and 
$\bm{D}$ is the diagonal weighting matrix containing the weights~$d^{[i]}$. 
In the covariance matrix update, $Y$ is the same as in Eq.~(\ref{eq: update_wml}). 
\change{
Here, $\lambda$ is a small positive variable and $\bm{I}$ is an identity matrix.
The introduce of $\lambda \bm{I}$ is for numerical stability when calculating the matrix inverse.}
A complete policy search for learning a Gaussian linear policy for MPC 
is given in Algorithm~\ref{algo: lineargaussianmpc}.
\change{A detailed derivation of the above solution is available in the Appendix.}

We use the Random Fourier Features~(RFF)~\cite{rajeswaran2017towards} as the featurizer
\begin{equation}
    \bm{\phi}(\bm{s})^{[i]} = \sin \left( \frac{\sum_i P_{ij} s^{j}}{v} + p^{[i]} \right)
\end{equation}
where each element~$P_{ij}$ is randomly sampled from~$\mathcal{N}(0, 1)$, $v$ is a 
bandwidth parameter, and $p$ is a random phase shift drawn from~$U[-\pi, \pi)$.
The bandwidth $v$ is the only parameter that has to be tuned. 
The RFF-based linear policy has been used to solve many benchmark continuous control tasks, including the OpenAI gym benchmarks~\cite{brockman2016gym}. 

\begin{algorithm}[!htp]
    \caption{{\bf Learning Gaussian Linear Policies for MPC} \label{algo: lineargaussianmpc}}
	\KwIn{ $\pi_{\bs{\theta}}(\mathbf{W}\phi(\mathbf{s}), \bs{\Sigma}), N, \text{MPC}, \mb{x}_0, \mb{p}$}
	\textbf{While not converged} \\
	\quad Sample observations: $\mathbf{s}^{[i]} \sim \rho(\mathbf{s})$ \\
	\quad Sample variables: $\bs{z}^{[i]} \sim \pi_{\bs{\theta}} \left(\mathbf{W}\phi(\mathbf{s}^{[i]}), \bs{\Sigma}\right)_{i=1...N}$\\
	\quad Sample trajectories: $\bs{\tau}^{[i]} = \text{MPC.solve}(\mb{x}_0, \bs{z}^{[i]}, \mb{p})$ \\
	\quad \textbf{Expectation:} \\
	\quad \quad $d^{[i]} = \exp{ \left\{ \beta R(\bs{\tau}^{[i]}) \right\}}$ \\
	\quad \textbf{Maximization:} \\
	\quad \quad $\bs{\theta}^{\ast} = \arg \max_{\bs{\theta}} \left\{ \sum_i d^{[i]} \log \pi_{\bs{\theta}} (\bs{z}^{[i]}| \mathbf{s}^{[i]}) \right\}$ \\
	\KwOut{Trained Policy $\pi_{\bs{\theta}^{\ast}}(\mathbf{W}^{\ast}\phi(\mathbf{s}), \bs{\Sigma}^{\ast})$}
\end{algorithm}

\subsection{Learning Neural Network Policies}
\change{Algorithm~\ref{algo: lineargaussianmpc} can optimize a linear policy 
using an episodic policy search method. 
Such episodic policy search by design is used for learning policies in multi-task settings, where distributions
over different tasks are well-defined. 
When controlling a robot in a highly dynamic environment, where the observations differ significantly from state to state,
we use a step-based policy search algorithm. 
Also, we aim to learn a complex neural network policy for selecting adaptive
decision variables and for processing relatively high dimensional observations. 
}
Such properties are potentially useful for the robot to adapt its behavior online in 
a highly dynamic environment.

We train the neural network policy by combining Algorithm~\ref{algo: gaussianmpc}
with supervised learning.
A complete algorithm of learning neural network policies for MPC is given
in Algorithm~\ref{algo: neuralmpc}.
We divide the learning process into two stages: 1) data collection, 2) policy learning.
In the data collection stage, we randomly initialize the robot in a state $\mb{x}_t$ and find 
the optimal decision variables $\mb{z}^{\ast}_t$ via Algorithm~\ref{algo: gaussianmpc}.
We aggregate the dataset by $\mathcal{D} \leftarrow \mathcal{D} \cup (\mb{o}_t, \mb{z}^{\ast}_t)$,
where $\mb{o}_t$ is the current observation of the robot.
A sequence of optimal control actions $\mb{u}^{\ast}_t$ are computed by solving the MPC optimization, 
given the current state $\mb{x}_t$ of the robot and the learned variable $\mb{z}^{\ast}_t$.
Only the first control command is applied to the robot; subsequently, the robot moves to the next state.
Incrementally, we collect a set of data that has diverse training pairs $(\mb{o}_t, \mb{z}^{\ast}_t)$
consisting of an observation~$\mb{o}_t$ as the neural network input and a ground-truth value~$\mb{z}^{\ast}_t$ as the output. 

It is important to note that at each simulation time step~$t$, we run Algorithm~\ref{algo: gaussianmpc}
to solve multiple MPC optimizations in order to find the optimal decision variable for the current state.
This step can be viewed as an online learning process in the simulator and is difficult to be run in real-time. 
During policy learning, we train the neural network by minimizing the mean-squared-error
between the labels $\mb{z}^{\ast}_t$ and the output of the network $f_{\mb{\Phi}}(\mb{o}_t)$,
using the standard stochastic gradient descent.
After training, the neural network policy is deployed together with the MPC 
to control the vehicle. 
Since the resulting controller contains a high-level policy and an MPC,
we name this controller High-MPC.
\begin{algorithm}[!htp]
    \caption{{\bf Learning Neural Network Policies for MPC}   
    \label{algo: neuralmpc}}
    \KwIn{ $f_{\bm{\theta}}, \mathcal{D}=\{ \}$ }
	\textbf{Data collection (repeat)} \\
	\quad  Randomly reset the system: $\mb{x}_t, \mb{o}_t, \mb{p}_t, t=0$\\
	\quad  While not done: \\
	\quad \quad $(\mb{z}_t = \bs{\mu}^{\ast}) \leftarrow$ \textbf{Algorithm~\ref{algo: gaussianmpc}} ($\mb{x}_0=\mb{x}_t, \mb{p}_t$) \\
	\quad \quad Data collection: $\mathcal{D} \leftarrow \mathcal{D} \cup \left\{\mb{o}_t, \mb{z}_t\right\}$ \\
	\quad \quad MPC optimization: $\mb{u}^{\ast}_t = \text{MPC.solve}(\mb{x}_t, \mb{z}_t, \mb{p}_t)$\\
	\quad \quad System transition: $\mb{x}_{t+1} \leftarrow f(\mb{x}_t, \mb{u}^{\ast}_t)$ \\
	\textbf{Policy learning} \\
	\quad $\bm{\theta}_\text{new} = \arg \min_{\bm{\theta}} \| f_{\bm{\theta}}(\mb{o}_t) - \mb{z}_t \|^2$ \\
	\KwOut{Learned deep high-level policy $f_{\bm{\theta}^{\ast}}$}
\end{algorithm}
\section{Flying A Quadrotor through dynamic gates}
\label{sec: application_preliminary}
We apply the proposed \textit{policy-search-for-MPC} framework to address a challenging 
problem towards agile drone flight in dynamic environments, 
which is learning to fly through dynamic gates. 
The ability to fly through fast-moving gates enables the drone
to traverse inside a dynamic environment, where the free space is changing 
rapidly.
It is a difficult task since it requires simultaneously planning an accurate trajectory that
passes through the center of moving gates and controlling the quadrotor to precisely follow the trajectory.

\subsubsection{Quadrotor Dynamics}
We model the quadrotor as a rigid body controlled by four motors.
The dynamics of the system can be written as:
\begin{align}
\mathbf{\dot{p}}_{WB}&=\mathbf{v}_{WB}&\mathbf{\dot{q}}_{WB} &= \frac{1}{2} \mathbf{\Lambda} ( \boldsymbol{\omega}_{B}) \cdot \mathbf{q}_{WB} \\
\mathbf{\dot{v}}_{WB}&=\mathbf{q}_{WB}\odot\mathbf{c}-\mathbf{g} & \boldsymbol{ \dot{\omega}_{B}} &= \mathbf{J}^{-1}(\boldsymbol{\eta} - \boldsymbol{\omega}_{B} \times \mathbf{J} \boldsymbol{\omega}_{B})  
\end{align}
where $\mathbf{p}_{WB}^q=[p_x^q, p_y^q, p_z^q]^T$ and $\mathbf{v}_{WB}^q=[v_x^q,v_y^q,v_z^q]^{T}$ are the position 
and the velocity vectors of the quadrotor in the world frame~${W}$.
We use a unit quaternion $\mathbf{q}_{WB}=[q_{w},q_{x},q_{y},q_{z}]^{T}$ to represent the orientation of the quadrotor
and use $\bs{\omega}_{B}= [\omega_x, \omega_y, \omega_z]^T$ to denote the body rates 
(roll, pitch, and yaw respectively) in the body frame~${B}$.
Here, $\mathbf{g}=[0, 0, -g_z]^{T}$ with $g_z=9.81 m/s^2$ is the gravity vector,
$\mathbf{J}$ is the inertia matrix, $\boldsymbol{\eta}$ is the three dimensional torque, 
and $\mathbf{\Lambda} (\bs{\omega}_{B})$ is a skew-symmetric matrix.
Finally, $\mb{c}=[0, 0, c]^T$ is the mass-normalized thrust vector. 
The full state of the quadrotor is defined as
$\mb{x}^q=[\mb{p}^q, \mb{q}^q, \mb{v}^q]$
(we omit subscript for clarity).

\subsubsection{Pendulum Dynamics}
We model the dynamic gate as a nonlinear pendulum.
We approximate the pendulum gate as a point mass that is suspended on 
a weightless and inextensible string
of length $L_{cm}$ from a fixed support whose position is $\bs{P}_{WP}=[x_f, y_f, z_f]$.
The pendulum is subject to three forces: the gravity, the tension force results from the 
string pulling upon the bob of the pendulum, and a damping force due to friction and air resistance. 
We approximate the damping force by $f_d=-b * \dot{\theta}$, where 
$\dot{\theta}$ is the angular velocity and $b \in \mathbb{R}_{+}$ is a damping factor. 
We consider the pendulum's motion in the $y-z$ plane, meaning the pendulum rotates about the $x-$axis. 
Dynamics of the rotational motion is described by the following differential equations
\begin{align}
    \label{eq: pendulum_rotational}
    \ddot{\theta}_x = -\left( m g_zL_\text{cm} \sin \theta_x / I + b\dot{\theta}_x\right),
    \ddot{\theta}_y = 0, 
    \ddot{\theta}_z = 0  
\end{align} where $\theta_x$ is the roll angle, and $I$ is the moment of inertia. 
Dynamics of the translational motion are given by 
\begin{align}
    \label{eq: pendulum_translational}
    \ddot{v}_x=0, \quad
    \ddot{v}_y = l\cos(\theta_x) \ddot{\theta}_x, \quad
    \ddot{v}_z = l\sin(\theta_x) \ddot{\theta}_x 
\end{align}
where $l$ is the distance between the gate center and the fixed point. 
For the computational convenience, we transfer the Euler angles into a unit quaternion~$\mb{q}_{WB}^{g}$
to represent the gate orientation.
The full state of the gate center with respect to the inertial frame
is represented using the state vector~$\mb{x}^g = [\mb{p}^{g}, \mb{q}^{g}, \mb{v}^{g}]$.


\subsection{Learning to Fly Through Dynamic Gates}
\textit{Trajectory Optimization and Cost Function:}
We formulate the problem of learning to fly through dynamic gates.
Our main goal is to find a trajectory 
that passes through the center of the moving gates. 
Such a trajectory optimization problem involves 
1) decide a sequence of traversal times at which should the dynamic gates be passed, 
2) given these traversal times, find a trajectory that passes these gates. 
Since the gates are moving quickly, 
the optimization faces a \textit{chicken-and-egg} dilemma, 
namely, without obtaining the traversal times, it cannot determine the gates' state from which the vehicle 
should fly through, or without the gates' state, it cannot decide the traversal times. 

We first make the assumption that 
a vector of desired traversal times~$\mathbf{t}=[t_1, \cdots, t_i]$ 
for each gate~$i$ is given, where~$0 < t_i < t_j, \text{if}~i < j~\text{and}~i, j \in [1, \cdots, n]$.  
Here, $n$ is the total number of moving gates. 
Since we know the current states and the dynamic model of the moving gates, we can predict the future trajectory~$\boldsymbol{\tau}_i = [\mathbf{x}^g_{0}, \cdots, \mathbf{x}^g_{t_i}]$ 
for each gate~$i$, where $\mathbf{x}^{g}$ is a state vector that consists of position~$\mathbf{p}^g$, linear velocity~$\mathbf{v}^g$, and orientation~~$\mathbf{q}^g$.
Therefore, we define a gate-pass cost $\mathcal{L}_\text{gate-pass}$ as the following quadratic cost
\begin{equation}
    \mathcal{L}_{\text{gate-pass}} = \sum_{h=1}^{H-1} (\mathbf{x}^{q}_{h} - \mathbf{x}^{g}_{t_i})^T \mathbf{Q}_{p}(\mathbf{x}^{q}_{h} - \mathbf{x}^{g}_{t_i}) \cdot p_h,
\end{equation}
where $\mathbf{Q}_{p}$ is a diagonal cost matrix and $p_h$ a Boolean variable defined as 
\begin{equation}
   p_h =\begin{cases}
    1, & \text{iff $h = \lfloor t_i / d_t \rfloor $},\\
    0, & \text{otherwise}.
  \end{cases}
\end{equation}
Minimizing this loss function encourages the discretized states $\mathbf{x}_h^q$ to stay as closer
as possible to the gate states~$\mathbf{x}_{t_i}^g$, but only at the given desired traversal time $t_i$
for the gate $i$. For other discretized states at $h\neq \lfloor t_i / d_t \rfloor$, the loss
has no effects since $\mathcal{L}_\text{gate-pass}=0$. 

However, such a cost formulation requires very accurate dynamic modeling for both
the quadrotor and the moving gates, since $\mathcal{L}_\text{gate-pass}$ only characterizes 
several sparse states that are close to time nodes of the given traversal time vector $\mathbf{t}$. 
In other words, the optimization treats the moving gates as a list of static waypoints
and takes them as soft constraints, without considering their dynamic motions. 
To counteract potential model errors and uncertainties, we define a gate-follow cost
\begin{align*}
    \mathcal{L}_\text{gate-follow} &= \sum_{h=1}^{H-1} (\mathbf{x}^{q}_{h} - \mathbf{x}^{g}_{t_i})^T \mathbf{Q}_{f}(\mathbf{x}^{q}_{h} - \mathbf{x}^{g}_{t_i}) \cdot w_h \cdot (1 - p_h), \\
    \text{where}, & \quad w_h = \exp{ (- \alpha \cdot (h \cdot d_t - t_i)^2 )} \cdot \gamma_i.
\end{align*}

Here, $\mathbf{Q}_{f}$ is a diagonal cost matrix, 
$\omega_h$ defines the exponential weights for following the gate's motion, $\alpha \in \mathbb{R}_{+}$ defines the temporal spread of the weight, and $\gamma_i \in \mathbb{R}_{+}$ specifies different weights for tracking different gates.
The gate-follow cost provides an intuitive motivation: plan a trajectory that follows the gate if the time difference between the current time $h*d_t$ and the desired traversal time $t_i$ is small; and does not follow the gate if the time difference is significant. 
In other words, it minimizes the difference between the quadrotor states and the gates' states gradually as the quadrotor approaches the gate. 

In addition, we define a terminal cost $\mathcal{L}_\text{terminal}$ and an action regularization cost $\mathcal{L}_\text{u}$:
\begin{align}
    \label{eq: mpc_terminal}
    \mathcal{L}_\text{terminal} & = (\mathbf{x}^{q}_{H} -\mathbf{x}^\text{goal})^T\mathbf{Q}_\text{goal}(\mathbf{x}^{q}_{H} -\mathbf{x}^\text{goal}) \\
    \label{eq: mpc_action}
    \mathcal{L}_\text{u} & = \sum_{h=1}^{H-1} (\mathbf{u}^{q}_{h} - \mathbf{u}_{r})^T \mathbf{Q}_\text{u} (\mathbf{u}^{q}_{h} - \mathbf{u}_{r})
\end{align}
where $\mathbf{x}^\text{goal}$ is a goal state for hovering and $\mathbf{u}_{r} = [g_z, 0, 0, 0]$.
Here, we use body-rate control $\mathbf{u}_{h} = [c, \omega_x, \omega_y, \omega_z]$ as the inputs.
The terminal cost encourages the quadrotor to fly toward 
a goal state $\mathbf{x}^\text{goal}$. 

In summary, we have the following optimization problem 
\begin{align*}
\min_{\boldsymbol{\tau}} \quad \mathcal{L}(\mb{x}_1, \mb{z}, \mb{r}) & =\mathcal{L}_\text{terminal}+\mathcal{L}_\text{gate-pass}+\mathcal{L}_\text{gate-follow}+\mathcal{L}_\text{u} \\
\text{s.t.:} & \quad \mathbf{u}_\text{min} \leq \mathbf{u} \leq \mathbf{u}_\text{max} \\
             & \mb{x}_{h+1} = \mb{x}_h + d_t \cdot \hat{f}(\mb{x}_h, \mb{u}_h), \quad \mb{x}_1 = \mb{x}_\text{init}
\end{align*}
where the $\bm{\tau}$ represents the generated trajectory that consists of 
the state vector $\mb{x}^q=[\mb{p}^q, \mb{q}^q, \mb{v}^q]$ and the control inputs $\mathbf{u}$.

We define a vector of high-level decision variables
$\mathbf{z} = [t_1, \gamma_1, \cdots, t_n, \gamma_n]$ that consists of the desired
traversal time $t_i$ and the desired weights $\gamma_i$ for each moving gates. 
Here, $n$ is the total number of gates. 
Given the decision variables~$\mathbf{z}$, the current state of the quadrotor~$\mathbf{x}_0$, 
the predicted future trajectories~$\mathbf{r}$ of all moving gates, 
we can solve the trajectory optimization problem and find the optimal trajectory 
for the quadrotor to fly through all moving gates
\begin{equation}
    \boldsymbol{\tau} = \text{MPC.solve}(\mathbf{z}, \mathbf{x}_0, \mathbf{p}). 
\end{equation}

\textit{Policy Search and Reward Function:}
The aforementioned optimization formulation relies on the assumption that the desired traversal time~$t_i$
for each individual gate is given. 
In addition, the gate-follow loss $\mathcal{L}_\text{gate-follow}$ requires additional variables~$\gamma_i$
to define the weights that are assigned to follow the gate motion in the $y-z$ plane. 
The high-level decision variables $\mathbf{z}$ are difficult to model or tune due 
to the dynamic properties of the task. 
Therefore, a key requirement for our MPC to solve the problem is to have both 
the desired traversal time and the desired weights in advance. 
A similar MPC formulation was discussed in~\cite{neunert2016fast}, where 
the time variable at which a static waypoint should be reached 
by a quadrotor is also unknown. 
They solved the problem by manually selecting the variable via trial-and-error.
In our case, the decision variables are much more difficult to select manually.

We solve the problem of finding optimal decision variables in MPC using policy search. 
We define a Euclidean distance reward function to evaluate the goodness 
of the trajectory generated by the optimization, 
\begin{equation}
    R(\boldsymbol{\tau} | \mathbf{z}) = - \sum_i^{n} \| \mathbf{p}^q_{h_i} -  \mathbf{p}^g_{h_i} \|_2 - \lambda t_i
\end{equation}
where $h_i = \lfloor t_i / d_t \rfloor$ is the time node at which the predicted quadrotor
trajectory intersects with the gate state.
This reward is a sparse signal that evaluates the ``performance" of the sampled $\mathbf{z}$---high rewards indicate
smaller traversal distance error and low rewards indicate large traversal distance error of the predicted trajectory. 
Hence, maximizing this reward signal leads to desired traversal time variables that
allow the optimization to find a trajectory that passes through the center of the gate.
Here, $\lambda t_i$ is a regularization term used for choosing smaller time variables. 

\section{Experiments}
We design our experiments to evaluate the proposed policy-search-for-MPC framework. 
Specifically, we aim to answer the following questions:
1) can we learn \textit{state-independent} policies for the optimization~(Section~\ref{sec: exp1}),
2) can we learn \textit{state-dependent} linear policies for solving the optimization 
under different contexts~(Section~\ref{sec: exp2}),
3) can we learn a neural network policy for adapting MPC on the fly~(Section~\ref{sec: exp3}),
4) and finally, what is the performance of our system in the real world~(Section~\ref{sec: exp4}). 

\subsection{Learning State-Independent Time Variables for Trajectory Optimization}
\label{sec: exp1}
The first problem is a single trajectory optimization problem, where the goal
is to find a safe trajectory that passes through several dynamic gates
for a given initial state. 
We define three dynamic gates that are modeled using the same pendulum dynamics and
are initialized at different positions.
We use a prediction time horizon of $t_H = \SI{3.0}{\second}$ and 
a discretize time step of $d_t=\SI{0.05}{\second}$ for trajectory optimization, 
it results in a total discretization of $H=40$ nodes. 
We use CasADi~\cite{andersson2019casadi} with
IPOP~\cite{wachter2006implementation} as the solver for the numerical optimization. 
We learn a vector of decision variables~$\mathbf{z}$ using Algorithm~\ref{algo: gaussianmpc},
where $\mathbf{z}$ is modeled as a high-level policy and is represented using
a Gaussian distribution~$\mathbf{z} \sim \pi_{\bm{\theta}}(\bm{\mu}, \bm{\Sigma})$. 

\begin{figure*}[!thp]
    \centering
    \includegraphics[width=1.0\linewidth]{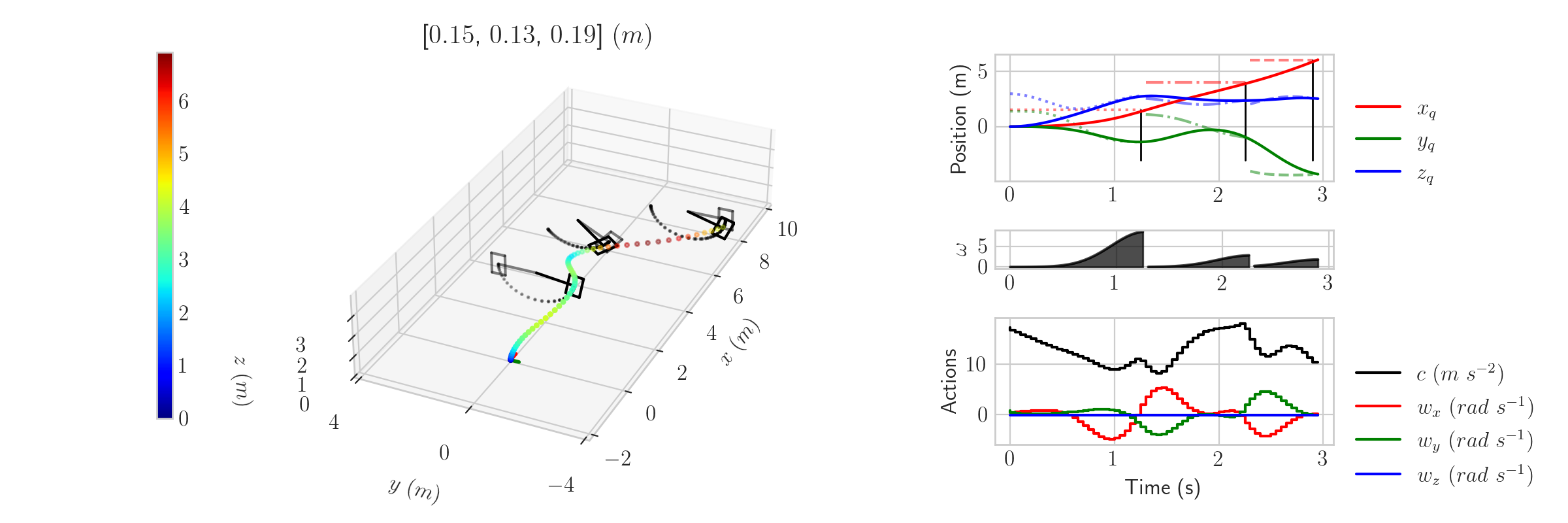}
    \caption{  
    \textbf{Left:} A planned trajectory for flying through 3 dynamic gates.
    The initial states of the moving gates are indicated by grey color.
    The quadrotor velocity (\SI{}{\meter\per\second}) is indicated by the color bar. 
    \textbf{Right:} \textit{Top:} The predicted positions of three moving gates (dashed line) and of the quadrotor (solid line).
    The learned traversal times for three gates (vertical line).
    \textit{Middle:} Learned weights~($\omega_h$) for the gate-following loss. 
    \textit{Bottom:} Control commands of the quadrotor.
    }
    \label{fig: evalGP}
\end{figure*}

Fig.~\ref{fig: evalGP} shows the predicted trajectory with the learned parameters.
The optimization successfully plans a trajectory that passes through the center of 
all moving gates at the given learned traversal time.
Besides, we achieve small traversal distance errors for all gates.
The distance errors for gates $(1, 2, 3)$ are $(\SI{0.13}{m}, \SI{0.15}{m}, \SI{0.30}{m})$ respectively. 
It is important to highlight that the predicted quadrotor position gradually follows the predicted gate's
center only when the quadrotor is close to the gate. 
Such a feature has crucial effects on real-world deployment since the dynamic modeling of the system is 
prone to error and the quadrotor has to follow the gate center when it is approaching the gate. 
Moreover, for the time stages that are far away from the desired traversal time,
the pendulum motion has less influence on the quadrotor,
leaving more extra control authority to counteract disturbance.

\begin{figure}
     \centering
     \includegraphics[width=0.5\textwidth]{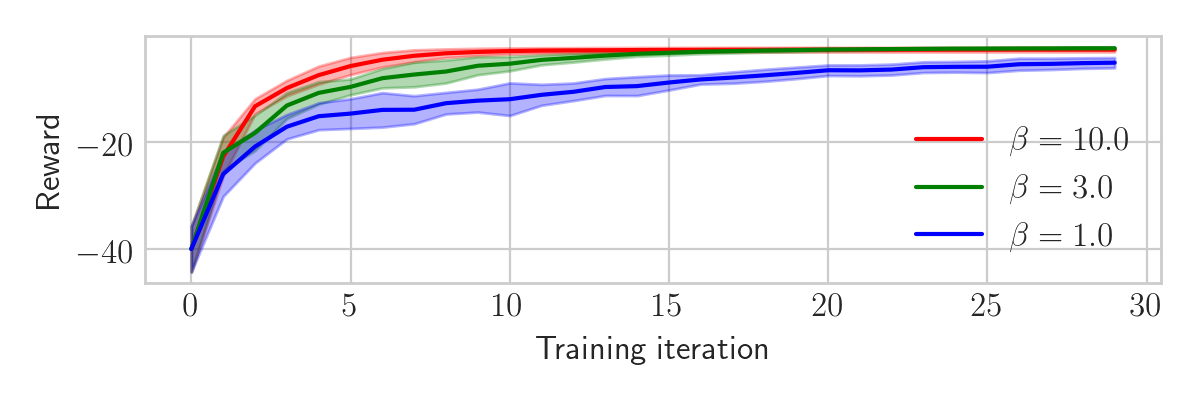}
     \caption{Learning curves of the Gaussian policy.
     Each curve is obtained using different temperature parameters $\beta$ for the policy training.
     We train six different randomly initialized policies for each temperature parameter. We compute the mean (solid line) and standard deviation (shadow region). }
     \label{fig: learningGP}
\end{figure}

Fig.~\ref{fig: learningGP} shows the learning progress of the Gaussian policy, 
which has randomly initialized weights (both mean and variance). 
The learning is data-efficient and stable as the policy converges after only
a few training iterations, e.g., 10 iterations for $\beta=10.0$.
We train the policy for 30 iterations to make sure that the policy is fully converged.
The policy update at each training step requires 30 samples, 
resulting in 900 MPC optimizations in total. 
Besides, a high temperature $\beta=10$ results in a fast (greedy) policy update 
while a low temperature can have slow convergence. 

\subsection{Learning State-Dependent Time Variables for 
Adaptive Trajectory Optimization Using A Linear Policy}
\label{sec: exp2}
For generalizing the learned high-level policies to different settings or contexts (denoted as a vector~$\mathbf{s}$),
we train a linear function approximator~$\mathbf{z} = f_{\mathbf{W}}(\mathbf{s})$. 
Specifically, we want to predict the decision variables conditioned on different initialization
of the moving gates. 
We characterize different settings using $\mathbf{s} = [\theta_x^1, \theta_x^2, \theta_x^3]$, where
$\theta_x^i, i=1, 2, 3$ are the initial angles of the gates about the $x$-axis and are randomly initialized.   
We represent the policy using a Gaussian linear model
$\mathbf{z}\sim\pi_{\bm{\theta}}(\mathbf{z}|\mathbf{W}\phi(\mathbf{s}),\bm{\Sigma})$~(Section~\ref{sec: pi_gplinear}),
where the policy parameters~$\bm{\theta}=[\mathbf{W},\bm{\Sigma}]$ are updated using Algorithm~\ref{algo: lineargaussianmpc}.
We use the RFF featurizer for $\phi(\mathbf{s})$, in which the feature bandwidth is specified as 0.1 
and the feature dimension is~40. 

Fig.~\ref{fig: learningGP} shows the learning progress of the Gaussian linear policy. 
Similar to training a Gaussian policy, the learning of a Gaussian linear policy is also 
very data-efficient and stable, thanks to the closed-form solution~(Eq.~(\ref{eq: sol_lineargp}))
for updating the policy parameters. 
Here, the policy update at each training step requires 300 samples,
resulting in 9000 MPC optimizations for learning a Gaussian linear policy. 

\begin{figure}[tp]
     \centering
     \includegraphics[width=0.5\textwidth]{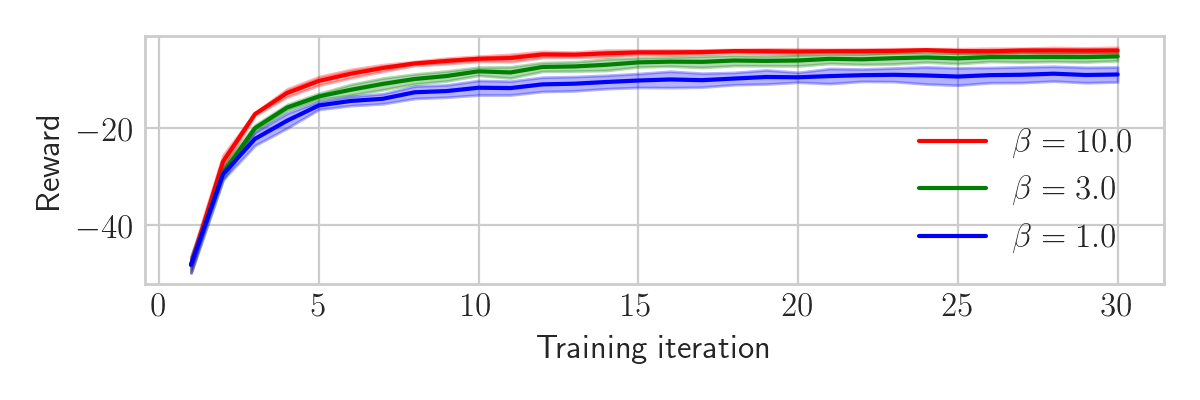}
     \caption{Learning curves of the Gaussian linear policy. 
     Each curve is obtained using different temperature parameters $\beta$ for the policy training.
     We train six different randomly initialized policies for each temperature parameter. 
     We compute the mean (solid line) and standard deviation (shadow region). }
     \label{fig: learningLinearGP}
\end{figure}

\begin{figure*}[tp]
    \centering
    \includegraphics[width=1.0\linewidth]{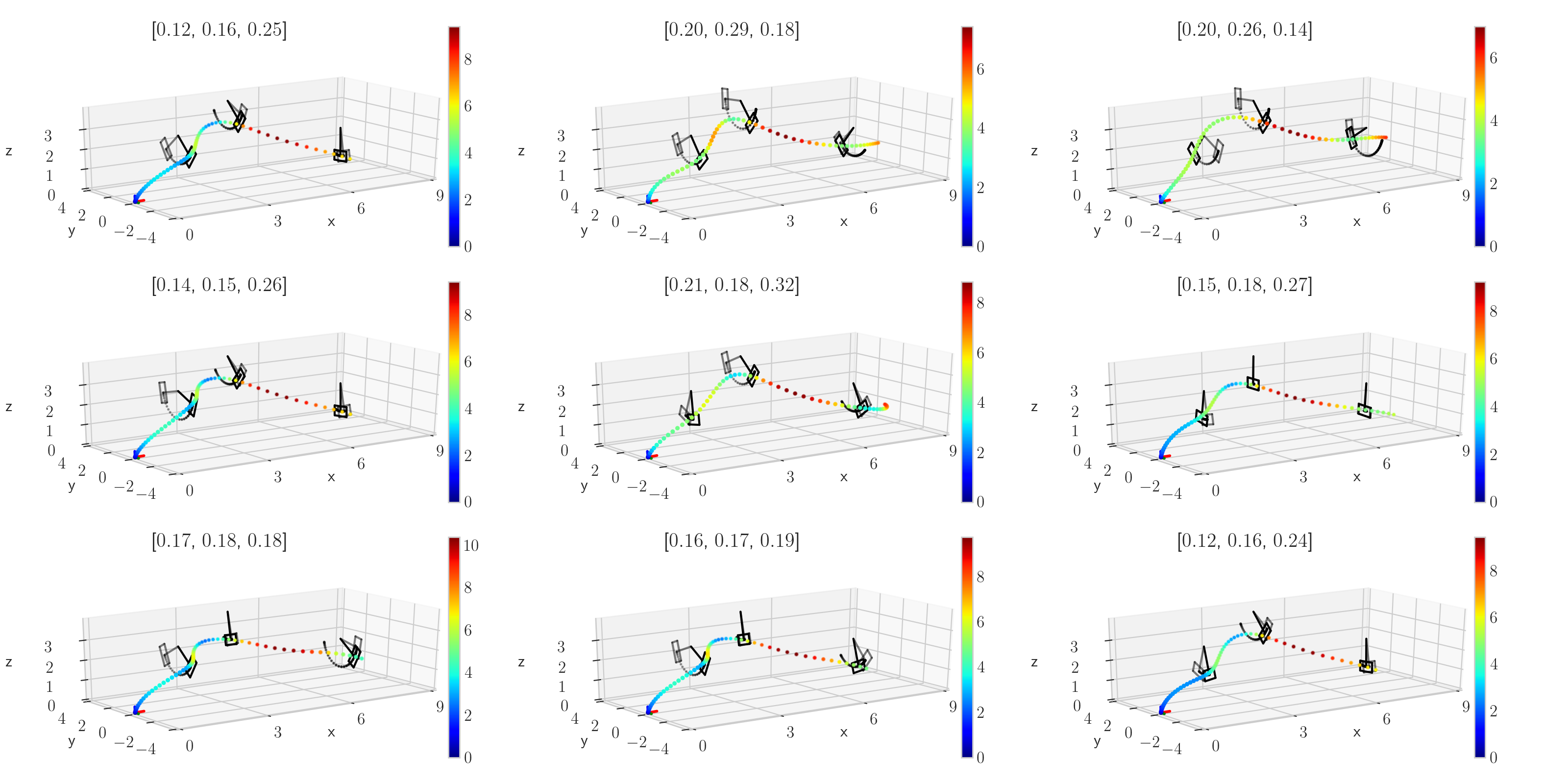}
    \caption{Evaluation of the trained linear high-level policy with randomly initialized pendulum states. 
    The generated quadrotor trajectories are colored by their speeds that are indicated using the color bars. 
    The traversal distance errors between the quadrotor center and the gate center at the desired traversal times are indicated in the figure title.}
    \label{fig: eval_LinearGP}
\end{figure*}

\change{
TABLE~\ref{tab: lineargaussian4mpc} shows the evaluation results.
The goal is to plan a trajectory for passing through 3 individual moving gates, where the initial states of the gates are randomly selected.
Given a planned trajectory, we compute the traversal distance from the quadrotor center to the gate center,
at the predicted time instances.
We run the experiment repeatedly 100 times and compute the mean and 
the standard deviation of the traversal distance errors. 
In addition, we define a success planning if the traversal distance errors for all gates are lower than $\SI{0.5}{\meter}$. 
We report the success rates for the 100 trials in TABLE~\ref{tab: lineargaussian4mpc}.
Note that G1, G2, and G3 represent each gate separately. 
Our approach outperforms two simple baselines: one uses randomly sampled time variables, and one uses a heuristic-based selection of the time variables.
Fig.~\ref{fig: eval_LinearGP} shows a visualization of 9 randomly sampled examples.}

\begin{table}[tp]
    \caption{Evaluation of the trained Linear High-Level Policy. }
    \label{tab: lineargaussian4mpc}
    \centering
    \begin{tabular}{c|c|ccc|ccc|ccc}
        \toprule
        \multirow{3}{4em}{Approach} 
            & Number   & \multicolumn{3}{c|}{Success Rates} & \multicolumn{3}{c|}{Mean} & \multicolumn{3}{c}{Standard Deviation} \\
            & of Gates &   &(\SI{}{\percent})&  &  &(\SI{}{\meter})&  &  &(\SI{}{\meter})& \\
             &      &  G1 & G2 & G3 & G1 & G2 & G3 & G1 & G2 & G3\\
        \midrule
        \multirow{1}{4em}{Random}  
            & 3 & 0 & 0 & 0 & 2.75 & 3.46 & 3.45 & 0.45 & 0.25 & 0.99 \\
        \midrule
        \multirow{1}{4em}{Heuristic}  
            & 3 & 100 & 0   & 0 & 0.14 & 1.98 & 1.14 & 0.02 & 0.50 & 0.29  \\
        \midrule
        \multirow{1}{4em}{ \textbf{Ours} }
        & \textbf{3} & \textbf{100} & \textbf{100} & \textbf{94} & \textbf{0.17} & \textbf{0.21} 
        & \textbf{0.30} & \textbf{0.04} & \textbf{0.06} & \textbf{0.15} \\
        \bottomrule
    \end{tabular}
\end{table}

\subsection{Learning Adaptive Variables for Dynamic Gates 
via Neural Network Policies}
\label{sec: exp3}
\change{
For learning a policy that is useful for the online parameters adaption
or compatible with high-dimensional sensory observations, we train a neural network policy.
The trained policy can hence be used for adaptively making high-level decisions
for the MPC at each control time step, resulting in a closed-loop controller. 
%
%
}
We use a Multilayer Perceptron~(MLP) as the policy representation.
In reality, since the environment is only partially observable, 
we can only observe one gate ahead.   
We define the observation at the time step $t$ as $\mathbf{o}_t = \mathbf{x}_t^q - \mathbf{x}_t^g$, 
which represents the difference between the quadrotor's current state~$\mathbf{x}_t^q$ 
and the next gate's state~$\mathbf{x}_t^g$.
The output of the neural network is the desired traversal time $\mathbf{z} = [t_1]$ for flying
through the next gate. 

We use Algorithm~\ref{algo: neuralmpc} to train the neural network policy in simulation,
in which it combines Algorithm~\ref{algo: gaussianmpc} with self-supervised learning. 
First, we reset the system in random states, meaning we use a randomly initialized position, velocity,
and orientation for the quadrotor and a random initial angle for the pendulum gate. 
We find the optimal traversal time $\mathbf{z}^{\ast}_t$ in this state using
the policy search algorithm~(Algorithm~\ref{algo: gaussianmpc}).
We create a training pair~$(\mathbf{o}_t, \mathbf{z}^{\ast}_t)$ by associating 
the current observation with the current optimal variables. 
Then, we simulate the quadrotor to the next state by 
solving the MPC optimization using the optimal traversal time $\mathbf{z}^{\ast}_t$
and apply the optimal control command to the simulated quadrotor.
We also integrate the pendulum dynamics to simulate the gate motion.

We repeat this process at each simulation time step until the quadrotor flies through
the gate or it reaches the maximum simulation steps. 
In total, we collect 40,000 samples, which takes a multi-core CPU several hours 
to collect the training data.
However, the total sampling time can be significantly reduced using parallel processing or multithreading. 
We implement a fully-connected MLP with two hidden layers of 32 units each and ReLU as the activation function. 
The training of network weights takes less than 10 minutes on a standard notebook
with an Nvidia Quadro P1000 graphics card.

\begin{figure}[tp]
     \centering
     \includegraphics[width=0.5\textwidth]{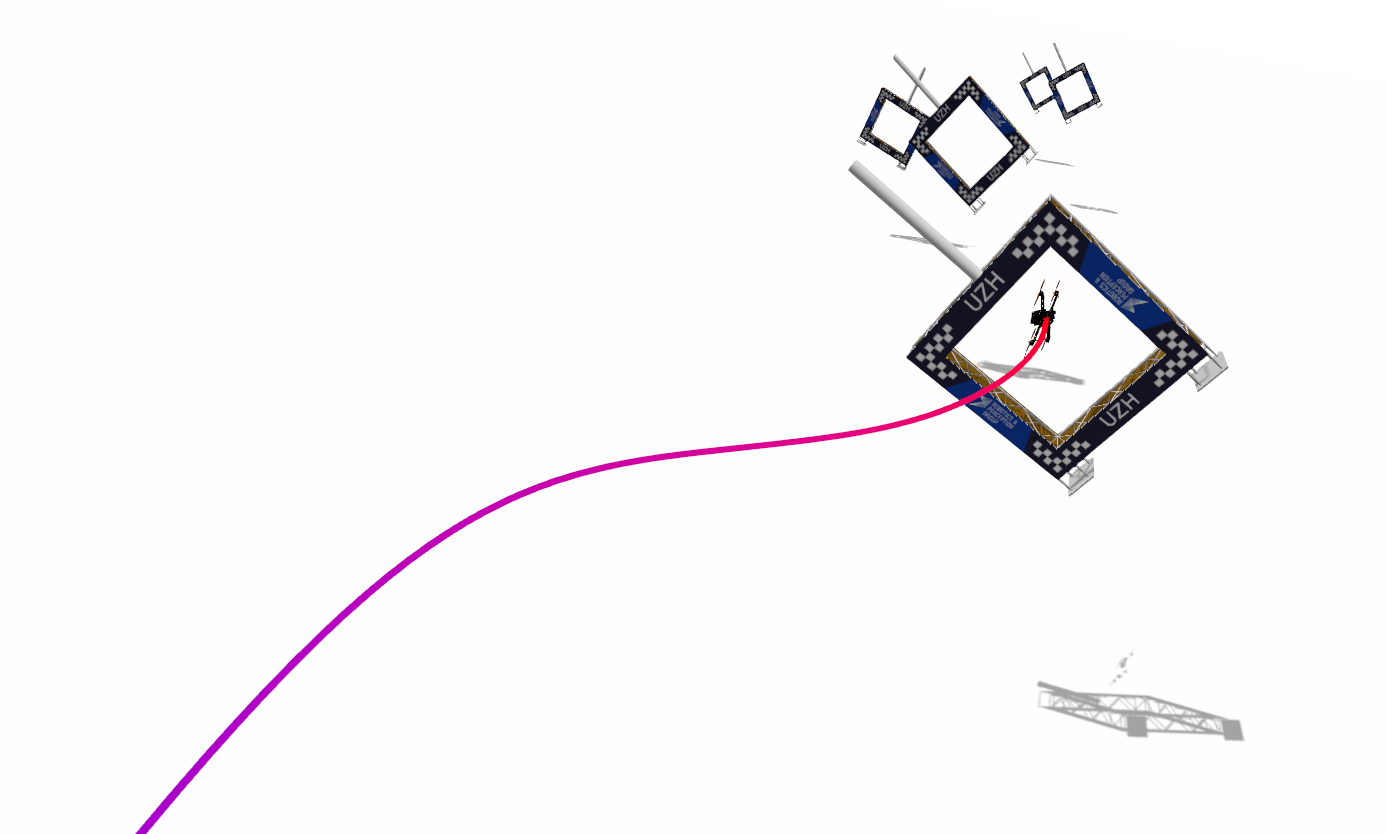}
     \caption{ A visualization of the dynamic racing environment in the Flightmare simulator~\cite{song2020flightmare}. }
     \label{fig: flightmare_race_track}
\end{figure}

We use an open-source quadrotor simulator called Flightmare~\cite{song2020flightmare} 
to simulate the race track and the quadrotor dynamics. 
Fig.~\ref{fig: flightmare_race_track} shows a visualization of the race track in Flightmare. 
We design a dynamic race track that contains 5 moving gates. 
All moving gates are modeled using the same pendulum dynamics and rotating around the $x$-axis.

The gates are attached to different fixed points and are initialized randomly by sampling 
its initial rotational angle~$\theta_x$ from a uniform distribution~$\theta_x \sim U(-\pi/2, \pi/2)$. 
Specifically, the gates are separated in the $x$-axis with a fixed position offset of $\delta_x$, 
where $\delta_x=p_x^{g, i} - p_x^{g, j}  \in [0.5, 1.0, 2.0, 3.0, 4.0, \cdots, 9.0]~(\SI{}{\meter})$ is the 
difference between two consecutive pendulum gates $i$ and $j$.
Besides, we initialize the quadrotor with different initial velocities of $v_x^q \in [0.0, 1.0, \cdots, 9.0]~(\SI{}{\meter\per\second}$)
in the forward direction.

We run 20 trials for each combination of the position offset and the initial quadrotor 
velocity $(\delta_x, v_x^q)$ and compute the success rate 
and the averaged traversal distance error. 
The averaged traversal distance error is computed using the Euclidean distance
from the quadrotor center  the gate.
We compute the success rate by defining a task as a success if all 5 traversal 
distance errors are less than $\SI{0.5}{m}$.

We compare our approach against two baselines: 
a standard MPC formulation, which does not have access to the desired traversal time,
and a fast motion primitive generator~\cite{mueller2015computationally}.
Since the desired time for flying through the next gate is not known beforehand, the desired traversal position is also a prior unknown.
We formulate a simple MPC optimization problem as the following
\begin{align*}
\min_{ \mb{u}, \mb{x} } \mathcal{L}  &= \mathcal{L}_\text{terminal} + \mathcal{L}_\text{u} + \sum_{h=1}^{H-1} ( \mathbf{x}^{q}_{h} - \mathbf{x}^{g}_{h})^T \mathbf{Q}(\mathbf{x}^{q}_{h} - \mathbf{x}^{g}_{h})\\
\text{s.t.:} & \quad \mb{u}_\text{min} \leq \mb{u} \leq \mb{u}_\text{max} \\
             & \mb{x}^{q}_{h+1} = \mb{x}^{q}_h + d_t \cdot \hat{f}(\mb{x}^{q}_h, \mb{u}_h), \quad \mb{x}^{q}_1 = \mb{x}_\text{init}
\end{align*}
where $\mathbf{x}^{g}_{h}$ is the predicted future state of the moving gate,
$\mathcal{L}_\text{terminal}$~(Eq.~\ref{eq: mpc_terminal}) is the terminal cost, 
and $\mathcal{L}_\text{u}$~(Eq.~\ref{eq: mpc_action}) is the action cost.
Here, $\mathbf{Q} = \text{diag}(0, 100, 100, 10, 10, 10, 10, 0, 10, 10)$ is a time-invariant diagonal cost matrix,
which is specified for tracking the gate motion in the $y$-axis and the $z$-axis.
Solving the above optimization problem results in a trajectory that follows the gate motion in the $y-z$ plane~(defined by the stage cost)
and reaches an end position located behind the gate~(defined by the terminal cost). 

As a second baseline, we use a minimum jerk trajectory~\cite{mueller2015computationally} together with a high-level
trajectory sampling scheme. 
We sample possible traversal trajectories by searching for the desired traversal time. 
We define a time interval of $T=[0, 3]$~\SI{}{\second} with a discretization time
step of $dt=$~\SI{0.1}{\second} for sampling. 
At each control time step, a total number of 30 trajectories are sampled. 
The start state of the trajectory is defined by the current quadrotor state.
The end state is partially defined, meaning some components are left free. 
The waypoint (the end position of the traversal trajectory) is calculated 
using a pendulum model of the moving gate and the sampled time. 
The end velocities and accelerations are defined as $\mathbf{v}=[\text{None}, 0, 0]$ and $\mathbf{a}=[\text{None}, \text{None}, \text{None}]$, 
which means that the end velocity $v_x$ in the forward direction ($x$-axis) and the accelerations are left free. 
We use zero velocities for the end state at the $y$-axis and $x$-axis to reduce the risk of having over-aggressive motion primitives. 
We reject trajectories that do not satisfy the input constraints and select
an optimal trajectory that has less traversal time.
%
\begin{figure*}[htp]
\centering
\includegraphics[width=1.0\textwidth]{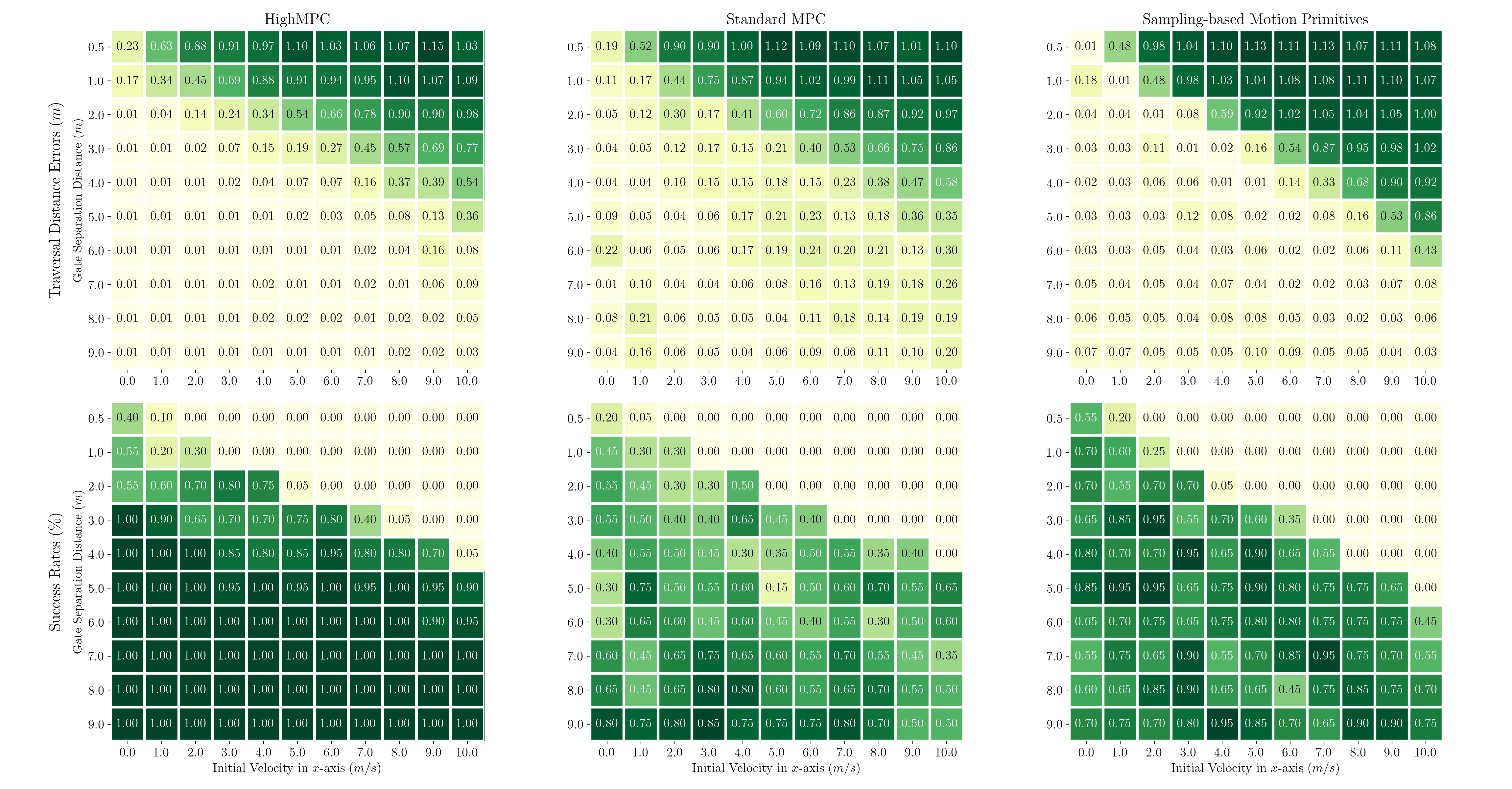}
\caption{
    Baseline comparison of flying through 5 moving gates using our High-MPC (left), 
    a standard MPC (middle), and a high-frequency motion primitive sampling method~\cite{mueller2015computationally} (right). 
    \change{Overall, our High-MPC outperforms both baselines in terms of gate traversal distance errors and success rates.}
}
 \label{fig: heatmap}
\end{figure*}

\change{
Fig.~\ref{fig: heatmap} shows the evaluation results. 
In summary, our approach achieves the highest success rates and smallest traversal distance errors,
particularly when the quadrotor has a small initial velocity and the 
separation distances among gates are large enough. 
When the quadrotor is initialized with a high forward velocity and the distance among gates are small,
Our approach is prone to failure of the task due to the physical constraint of the platform. 
For example, the quadrotor does not have enough control authority
to break immediately.

The standard MPC has achieved lower success rates and larger traversal distance errors.
This is due to the cost function formulation in the MPC optimization, in which the optimization does not have access to the desired traversal time, and thus, has to minimize the distance between the quadrotor and the gate center in all optimization states. 
Such an optimization scheme results in very aggressive trajectories and large control inputs~\cite{Yunlong2020learning}. 
Therefore, it is very important to obtain the traversal time prior to the optimization
and update the time at each control time adaptively for the closed-loop control. 
Optimizing the time jointly with other optimization variables is also possible,
but might result in a complicated optimization problem that is difficult to solve in real-time.

The sampling-based motion primitives method shows a competitive performance to our method. 
There are multiple challenges when applying the sampling-based method to our task.
First, the motion primitive is defined for state-to-state transition, in which the end velocities are difficult to be specified in advance. 
For example, a small velocity in the flight direction can result in a slow forward motion while a high velocity can lead to aggressive flight and gate overshooting. 
Second, the final performance of a task highly depends on the high-level planner, which is used for selecting the optimal motion primitive. 
Designing such a high-level planner is a challenging task, in which prior research
generally uses heuristic-based search~\cite{mueller2015computationally, zhou2019robust, liu2018search}, which could produce suboptimal solutions. }

\subsection{Real-world Deployment}
\label{sec: exp4}
\begin{figure}[t]
     \centering
     \includegraphics[width=0.5\textwidth]{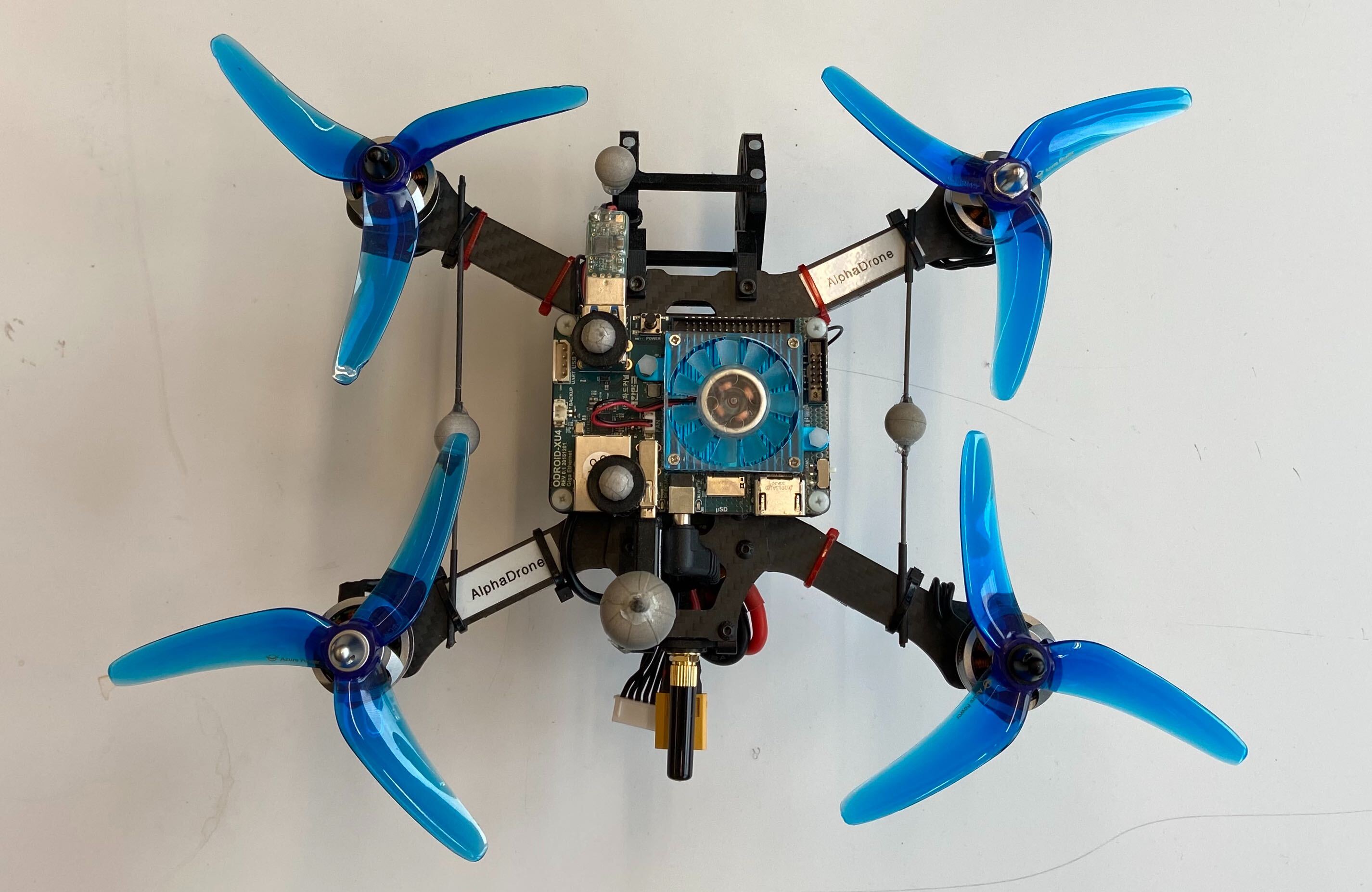}
     \caption{A racing drone used for the real-world experiment.}
     \label{fig: racing_drone}
\end{figure}

To test the robustness and the real-time performance of our system,
we deploy our approach in the real world with a physical racing drone.
The drone is built from off-the-shelf components used for first-person-view
racing~(Fig.~\ref{fig: racing_drone}). 
The drone features a carbon frame with stiff 5-inch propellers, a Lumenier flight controller,
an Odroid XU4 single board computer, and a Laird RM024 radio module for receiving control commands.
The platform weights \SI{0.775}{\kg}. 
We design a pendulum gate that comprises a straight steel stick and a wooden loop. 
The stick has a weight of \SI{0.3}{\kg} and a length of \SI{1.0}{\meter}.
The wooden loop has a weight of \SI{0.46}{\kg} and a radius of \SI{0.45}{\meter}.
We attach the pendulum gate to a fixed stand. 
The task goal is to control the quadrotor to fly through the pendulum gate
and hover it at a goal position located behind the gate. 

All the presented flight experiments were conducted with an OptiTrack motion capture system. 
We use Extended Kalman Filters~(EKF) for estimating both 
the quadrotor state and the pendulum state using observations from the OptiTrack. 
The state estimation runs at \SI{200}{\hertz}.
We use ACADO~\cite{Houska2011a} for the MPC optimization and qpOASES as a solver in order to achieve
real-time control performance of the quadrotor. 
As discretization step, we chose $d_t=\SI{0.1}{\second}$ with a prediction 
time horizon of $t_T = \SI{2.0}{\second}$.
The control command solved by the MPC with a desired traversal time variable 
obtained from the neural network high-level policy are updated at \SI{50}{\hertz}.
Our approach can achieve real-time control and is computationally efficient. 
The computational time for solving an MPC optimization at each control
time step is on average less than \SI{5}{\milli\second} and for the neural network
inference is on average less than \SI{2}{\milli\second} (without TensorRT optimization).

\begin{figure*}[htp]
  \centering
  \includegraphics[width=1.0\textwidth]{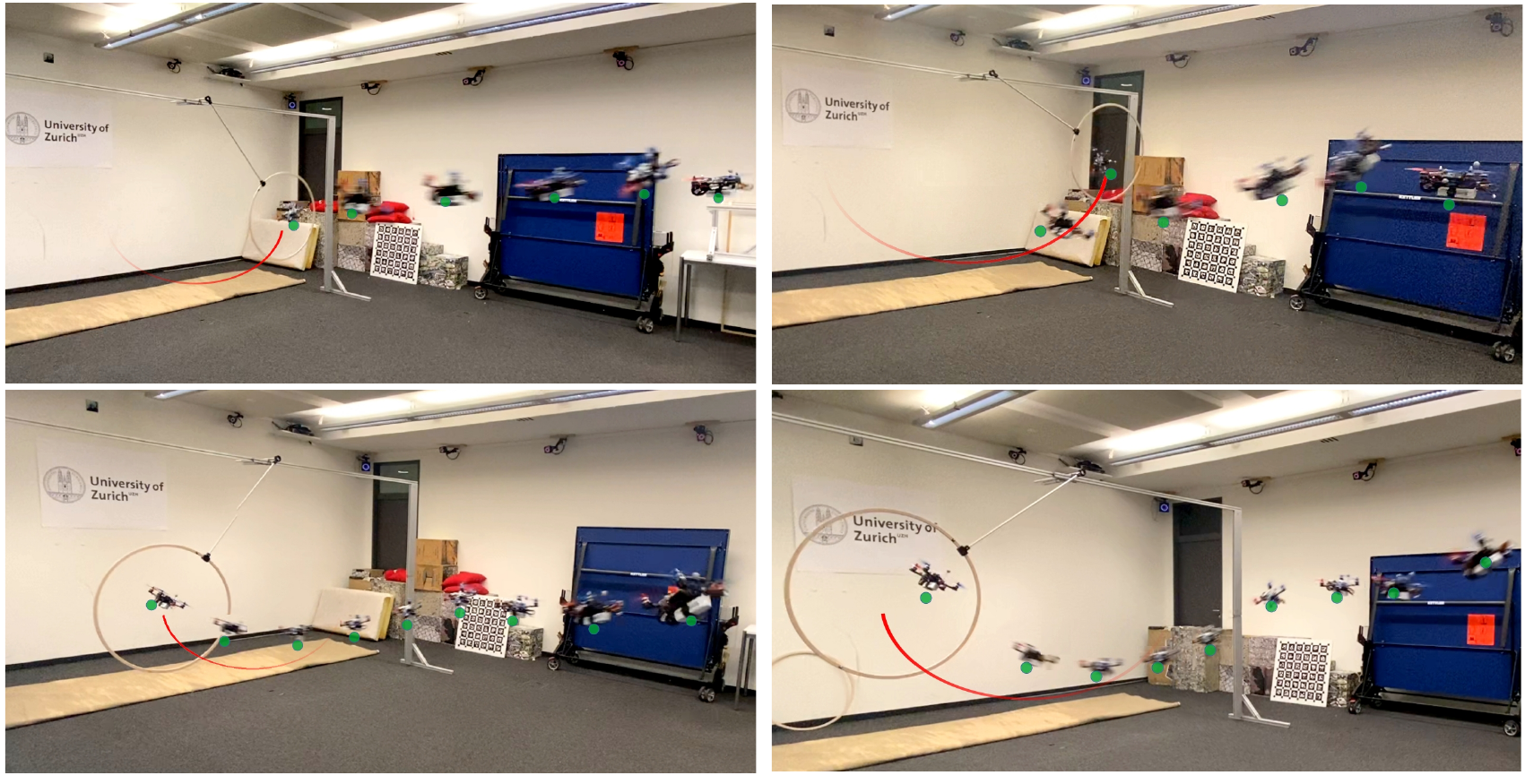}
  \caption{Controlling a quadrotor to fly through a dynamic gate using High-MPC under random initializations.}
  \label{fig: fly_trajectory}
\end{figure*}

We set up the experiment by putting the pendulum gate at a random initial angle and then dropping it. 
The pendulum gate swings back and forth freely with a periodic motion. 
The period of the pendulum decreases over time due to friction and air drag, which are difficult to model precisely. 
We approximate the pendulum dynamics using Eq.~(\ref{eq: pendulum_rotational}) \& Eq.~(\ref{eq: pendulum_translational})
with a roughly estimated damping factor $b=0.2$.
We use a simple dynamic model for predicting the future trajectory of the gate. 
We place our drone at a random position in front of the gate. 
Fig.~\ref{fig: fly_trajectory} shows four different trials of the real world experiment
conducted in a confined environment.
As a result, our approach successfully controls the quadrotor to fly through the fast-moving gate. 

\begin{figure*}[htp]
  \centering
  \includegraphics[width=1.0\textwidth]{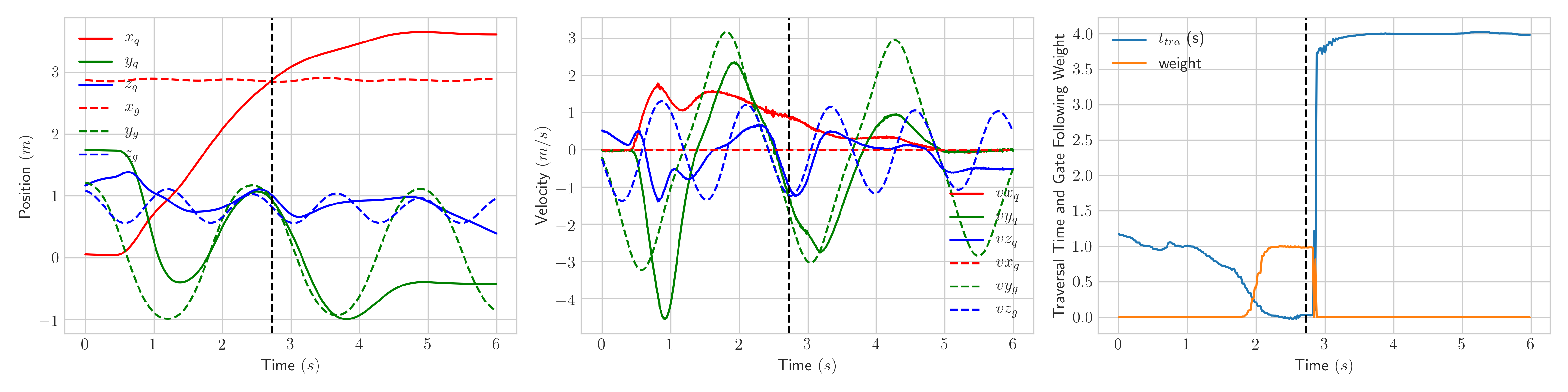}
  \caption{Trajectories of the quadrotor (solid line) and the pendulum gate (dashed line). 
  The plot on the right-hand side shows the predicted traversal time variable and the weight for minimizing the gate-following cost. 
  The vertical dashed line indicates the traversal time.
  }
  \label{fig: real_world_fly}
\end{figure*}

Fig.~\ref{fig: real_world_fly} shows a comparison of the executed trajectories between the quadrotor
and the moving gate.
The vertical black dashed line indicates the traversal moment, at which the quadrotor flies
through the gate. 
The quadrotor follows the gate's motion in both the $y$-axis (blue) and the $z$-axis when it
approaches to the gate. 
Besides, the desired traversal time~$t_\text{tra}$ predicted by the neural network 
together with the corresponding gate-following weighting variable~$\omega$ is plotted on the right-hand side.
Both $t_\text{tra}$ and $\omega$ are used by the MPC for simultaneous planning a trajectory and
controlling the vehicle. 
When $t_\text{tra}$ is close to zero, the weighting variable increases to a maximum value $\omega \approx 1$, 
which indicates that the desired traversal time is now and the quadrotor 
should try to follow the periodic motion in the $y-z$ plane.
By contrast, when $t_\text{tra}$ is very large (e.g.,  $t_\text{tra}=4.0$), 
the weighting variable decreases to a minimum value $\omega \approx 0$, 
meaning that there is no need to follow the gate since the gate is either far away or has been passed already.  
\change{
Note that $t_\text{tra}$ is not the prediction horizon in the MPC optimization.}
%

\subsection{Connections with Prior Trajectory Planning Methods}
In general, existing works on quadrotor trajectory planning (or kinodynamic motion planning) in dynamic environments 
can be divided into sampling-based and optimization-based approaches.
Sampling-based algorithms~\cite{elbanhawi2014sampling}, such as RRT$^{\ast}$~\cite{karaman2011sampling, webb2013kinodynamic}, 
are provably optimal in the limit of infinite samples. 
To achieve real-time replanning, sampling-based algorithms leverage
efficient motion primitive generators to provide a closed-form solution for state-to-state trajectories, e.g., the minimum jerk~\cite{mueller2015computationally} or minimum snap~\cite{mellinger2011minimum} trajectories. 
This line of work manifests a significant computational advantage and real-time planning
performance, however, relies on simplified dynamics or differentially flat dynamics of a quadrotor.
Moreover, they need to relax the single actuator constraints to limit the per-axis acceleration,
which might render planned trajectories conservative. 

Optimization-based approaches overcome these limitations by enforcing the system dynamics and thrust limits as constraints.
They use discrete-time state-space representations for the trajectory, and  
can handle nonlinear dynamics and different constraints.
For passing through dynamic gates, the allocation of the traversal times for the moving waypoints 
is a priori unknown, rendering the problem formulation complicated. 
\change{
State-of-the-art approaches address similar problems for passing through static waypoints using either heuristic search~\cite{neunert2016fast} or formulating a complicated optimization problem using complementary progress constraints~\cite{foehn2020cpc}.
}

%

\section{Discussion}
\label{sec: discussion}


\subsection{Choice of Policy Representations}
\change{
In Section~\ref{sec: ps4mpc}, we have presented algorithms for training three different policies, including Gaussian policy, Gaussian linear policy, and neural network policy. 
Our formulation is general and is designed to be suitable for a wide variety of robotic tasks.

The Gaussian policy finds state-independent decision variables for a single
trajectory optimization, namely, the learned decision variables are fixed parameters. 
In practice, many MPC applications require manually tuning of the hyperparameters, such as 
the prediction horizon. 
The Gaussian policy is useful for automatically selecting those hyperparameters. 

The Gaussian linear policy learns state-dependent policies via episodic policy search
and function approximations.
Such policy representations are useful for generalizing
robot skills to multiple contexts: a lower-level MPC, parameterized via the parameters~$Z$, 
controls the robot for a given context~$s$ and a high-level policy~$\pi(Z | s)$ generalizes among different contexts~$s$.
In the context of reinforcement learning, it is generally referred to as contextual policy search~\cite{deisenroth2013survey, kupcsik2013data}. 
Hence, the Gaussian linear policy has the advantage of efficiently learning linear function representations (with nonlinear kernel features) using a small number of samples.
The neural network policy learns state-dependent policies in a step-based setting, 
where the policy $\pi(Z_t | o_t)$ needs to adapt its decision variables $Z_t$ 
based on the observation $o_t$ at each control time step $t$. 
The above-mentioned episodic policy search cannot be used directly because the 
step-based policy search uses different exploration strategies and 
relies on different policy evaluation methods.
Instead, we propose to combine Algorithm~\ref{algo: gaussianmpc} with a self-supervised learning
scheme for the step-based setting. 
We show the resulting algorithm can be used for learning a complex neural network policy. 
The neural network policy is more helpful for adapting decision variables online
or potentially for processing high-dimensional observations.}

\subsection{Design of the Loss Function and the Reward Function}
The loss function design in the MPC optimization involves generating desired control commands for the robot such that the predicted future states match predefined high-level goals. 
Our approach allows more flexible and automatic loss function
design: by taking the MPC as a parameterized controller and using policy search for automatically selecting the desired parameters. 
The reward function design for policy search has very flexible formulations, such as quadratic, sparse, or exponential. 

\subsection{Limitations}
First, learning neural networks is computationally expensive
and data-hungry.
During data collection, the proposed self-supervised policy search~(see Algorithm~\ref{algo: neuralmpc}) requires running Algorithm~\ref{algo: gaussianmpc} in the loop, which is an expensive process since it needs to solve multiple MPC optimizations for each state. 
\change{
Second, Algorithm~\ref{algo: neuralmpc} was designed for learning neural networks
for both online decision making and handling high-dimensional observations.
Our experiments did not exploit the full potential of using deep neural networks for processing high-dimensional observations, such as images. 
Finally, our experiment results are based on accurate state estimation with low latency, which is general not the case when using onboard sensing and computing.
It is also crucial to consider many factors for real-world scenarios,
such as noisy state estimation and system delays.
}

\section{Conclusion}
\change{
This paper proposed a novel framework that unifies the advantages of both probabilistic policy search and MPC.
Our approach improves over the standard MPC formulation by augmenting the MPC controller
with learned high-level policies that can automatically choose hard-to-optimize decision variables.
Our framework allows a versatile design of different policy representations, ranging from state-independent Gaussian policies to complex neural networks.

As a second contribution, we addressed a challenging problem in agile drone flight:
controlling a quadrotor to fly through fast-moving gates. 
We successfully demonstrated the effectiveness of our approach in both simulation and the real world. 
To the best of our knowledge, our approach is the first to attempt to address this problem and, hence, can serve as an important baseline for future work. 

We release the source code of learning high-level policies for MPC 
and provide additional theoretical derivations in the Appendix. 
Future work concerns improving the sample efficiency of the high-level policy training using more advanced policy search methods, such as relative entropy policy search~\cite{peters2010relative}. 
On the application side, applying the proposed framework to other robotic tasks, such as addressing more complex trajectory optimization problems or using a high-level policy for dealing with model errors, are promising avenues for future research.}




\bibliographystyle{IEEEtran}
\bibliography{ref}

\section*{ACKNOWLEDGMENT}
We thank Thomas L\"angle, Roberto Tazzari, Manuel Sutter, Elia Kaufmann, Antonio Loquercio, 
Philipp Foehn, Angel Romero, and Sihao Sun for their help or the valuable discussions. 
\appendix

We provide detailed derivations for both updating the Gaussian policy and 
the Gaussian linear policy using weighted maximum likelihood. 
\section{}

\subsection{Derivation of Algorithm 1}
The objective of maximizing a Gaussian policy in Algorithm 1 is defined as
\begin{equation*}
    \bs{\theta}^{\ast} = \arg \max_{\bs{\theta}} \left\{ \sum_i d^{[i]} \log \pi_{\bs{\theta}} (\bs{z}^{[i]} ) \right\}
\end{equation*}
where the log-likelihood of the Gaussian policy is given by 
\begin{align*}
& \log \pi_{\bs{\theta}} (\bs{z} |\bs{\theta}) \\
& \quad =  \log \mathcal{N} (\bs{z} | \bs{\mu}, \bs{\Sigma} )\\
& \quad = \log \frac{ \exp{ \left( - \frac{1}{2} (\bs{z} -\bs{\mu})^T \bs{\Sigma}^{-1} (\bs{z} -\bs{\mu}) \right) }}{ \sqrt{ (2 \pi )^k |\bs{\Sigma}| } } \\
& \quad = - \frac{k}{2}\log (2 \pi) - \frac{1}{2} \log |\bs{\Sigma} | - \frac{1}{2} (\bs{z} -\bs{\mu})^T \bs{\Sigma}^{-1} (\bs{z} -\bs{\mu})\\
\end{align*}

In order to find the $\bs{\theta}$ that maximizes the reward, we take the derivative to the policy parameters $\bs{\theta}= [\bs{\mu}, \bs{\Sigma}]$, separately, and set the gradients to zero.
We first compute the solution for updating the mean~$\bs{\mu}$
\begin{align*}
& \nabla_{\bs{\mu}}  \sum_i d^{[i]} \log \pi_{\bs{\theta}} (\bs{z}^{[i]} )  \\
& \quad = \sum_i d^{[i]} \nabla_{\bs{\mu}} \log \pi_{\bs{\theta}} (\bs{z}^{[i]} )  \\
& \quad = \sum_i d^{[i]} \nabla_{\bs{\mu}} \Big( - \frac{k}{2}\log (2 \pi) -  \frac{1}{2}\log |\bs{\Sigma} |  \\
& \quad - \frac{1}{2} (\bs{z}^{[i]} -\bs{\mu})^T \bs{\Sigma}^{-1} (\bs{z}^{[i]} -\bs{\mu}) \Big) \\
& \quad = \sum_i d^{[i]}  \left(-\nabla_{\bs{\mu}} \frac{1}{2} (\bs{z}^{[i]} -\bs{\mu})^T \bs{\Sigma}^{-1} (\bs{z}^{[i]} -\bs{\mu}) \right) \\ 
& \quad = + \sum_i d^{[i]} (\bs{z}^{[i]} -\bs{\mu})^T \bs{\Sigma}^{-1} = \mathbf{0}.
\end{align*}
By solving for~$\bs{\mu}$, we obtain 
\begin{equation*}
\bs{\mu}_\text{new} = \frac{\sum_{i=1}^N d^{[i]} \bs{z}^{[i]}}{\sum_{i=1}^N d^{[i]}}  
\end{equation*}

Second, we compute the solution for the covariance matrix~$\bs{\Sigma}$,

\begin{align*}
& \nabla_{\bs{\Sigma}} \sum_i d^{[i]} \log \pi_{\bs{\theta}} (\bs{z}^{[i]} )  \\
& \quad = \sum_i d^{[i]} \nabla_{\bs{\Sigma}} \Big( - \frac{k}{2}\log (2 \pi) -  \frac{1}{2}\log |\bs{\Sigma} |  \\
& \quad\quad -   \frac{1}{2} (\bs{z}^{[i]} -\bs{\mu})^T \bs{\Sigma}^{-1} (\bs{z}^{[i]} -\bs{\mu}) \Big) \\
& \quad = \sum_i d^{[i]} (-\frac{1}{2}\bs{\Sigma}^{-1} + \frac{1}{2}\bs{\Sigma}^{-1} (\bs{z}^{[i]} -\bs{\mu})(\bs{z}^{[i]} -\bs{\mu})^{T}  \bs{\Sigma}^{-1} ) \\
& \quad = -\frac{1}{2} \bs{\Sigma}^{-1} \sum_i d^{[i]} + \\ 
& \quad\quad \frac{1}{2} \bs{\Sigma}^{-1} \left( \sum_i d^{[i]}(\bs{z}^{[i]} -\bs{\mu})(\bs{z}^{[i]} -\bs{\mu})^{T} \right) \bs{\Sigma}^{-1} \\
& \quad = \mathbf{0}.
\end{align*}

By solving for~$\bs{\Sigma}$, we obtain
\begin{equation*}
\bs{\Sigma}_\text{new} = \frac{\sum_{i=1}^N d^{[i]}(\mb{z}^{[i]}-\bs{\mu})(\mb{z}^{[i]}-\bs{\mu})^T}{\sum_i d^{[i]}} .
\end{equation*}
Note that in Eq.~(\ref{eq: update_wml}), we use
\begin{equation*}
Y = \frac{\left(\sum_{i=1}^{N} d^{[i]} \right)^2 - \sum_{i=1}^N (d^{[i]})^2 }{\sum_{i=1}^N d^{[i]}} 
\end{equation*}
as the demonimator to obtain an unbiased estimate of the covariance. 

\subsection{Derivation of Algorithm 2}
\begin{equation*}
    \bs{\theta}^{\ast} = \arg \max_{\bs{\theta}} \left\{ \sum_i d^{[i]} \log \pi_{\bs{\theta}} (\bs{z}^{[i]} | \mathbf{s}^{[i]} \ ) \right\}.
\end{equation*}
where the log-likelihood of the Gaussian policy is given by 
\begin{align*}
& \log \pi_{\bs{\theta}} (\bs{z} |\mathbf{s}; \bs{\theta}) \\
& \quad =  \log \mathcal{N} (\bs{z} | \mathbf{W}\bm{\phi}(\mathbf{s}), \bs{\Sigma} )\\
& \quad = \log \frac{ \exp{ \left( - \frac{1}{2} (\bs{z} - \mathbf{W}\bm{\phi}(\mathbf{s}))^T \bs{\Sigma}^{-1} (\bs{z} -\mathbf{W}\bm{\phi}(\mathbf{s})) \right) }}{ \sqrt{ (2 \pi )^k |\bs{\Sigma}| } } \\
& \quad = - \frac{k}{2}\log (2 \pi) - \frac{1}{2} \log |\bs{\Sigma} | \\
& \quad\quad  - \frac{1}{2} (\bs{z} -\mathbf{W}\bm{\phi}(\mathbf{s}) )^T \bs{\Sigma}^{-1} (\bs{z} -\mathbf{W}\bm{\phi}(\mathbf{s}))\\
\end{align*}

Similar to the previous derivation, in order to find the $\bs{\theta}$ maximizing the reward, 
we take the derivative to the policy parameters
$\bs{\theta}= [\bm{W}, \bs{\Sigma}]$, separately, and set the gradients to zero.
We first compute the solution for the parameters~$\bm{W}$ that are used for approximating the mean,
\begin{align*}
& \nabla_{\mathbf{W}} \sum_i d^{[i]} \log \pi_{\bs{\theta}} (\bs{z}^{[i]}| \mathbf{s}^{[i]})  \\
& \quad =  \frac{1}{2} \sum_i d^{[i]} \nabla_{\mathbf{W}} (\bs{z}^{[i]} -\mathbf{W}\bm{\phi}(\mathbf{s}) )^T \bs{\Sigma}^{-1} (\bs{z}^{[i]} -\mathbf{W}\bm{\phi}(\mathbf{s})) \\ 
& \quad = \bs{\Sigma}^{-1}  \sum_i d^{[i]} (\bs{z}^{[i]} -\mathbf{W}\bm{\phi}(\mathbf{s}))\bm{\phi}(\mathbf{s})^T  = \mathbf{0} \\
\Rightarrow  & \sum_i d^{[i]} \bs{z}^{[i]} \bm{\phi}(\mathbf{s})^T  =\sum_i d^{[i]}\mathbf{W}\bm{\phi}(\mathbf{s}) \bm{\phi}(\mathbf{s})^T.
\end{align*}

We can rewrite above equation in a matrix form
\begin{align*}
 & \mathbf{W}^T  \bm{\Phi}^{T} \mathbf{D}\bm{\Phi} = \mathbf{Z}^T \mathbf{D} \bm{\Phi} \\
\Rightarrow & \bm{W} =\left(\bm{\Phi}^{T} \bm{D} \bm{\Phi} \right)^{-1} \bm{\Phi}^{T} \bm{D} \bm{Z} \\
\end{align*}
where $\bm{\Phi}=[\phi(\bm{s}^{[i]} ) , \cdots, \phi(\bm{s}^{[N]})]$ is a matrix that 
contains converted feature vectors for all sampled observations~$\bm{s}$ and 
$\bm{D}$ is the diagonal weighting matrix containing the weights~$d^{[i]}$. 
Here, $\bm{Z}$ contains the sampled parameters $[ \bs{z}^{i}, \cdots, \bs{z}^{N}]$. Note that, in Eq.~(\ref{eq: sol_lineargp}), the introduce of $\lambda\mathbf{I}$ is for numerical stability.

Second, we compute the solution for the covariance matrix~$\bs{\Sigma}$,
\begin{align*}
&\nabla_{\bm{\Sigma}} \sum_i d^{[i]} \log \pi_{\bs{\theta}} (\bs{z}^{[i]}| \mathbf{s}^{[i]})  \\
&\quad =  -\frac{1}{2} \sum_i d^{[i]} \nabla_{\bm{\Sigma}} ( \log  |\bs{\Sigma} |  \\ 
& \quad + (\bs{z}^{[i]} -\mathbf{W}\bm{\phi}(\mathbf{s}) )^T \bs{\Sigma}^{-1} (\bs{z}^{[i]} -\mathbf{W}\bm{\phi}(\mathbf{s}))  ) \\ 
& \quad = -\frac{1}{2} \bs{\Sigma}^{-1} \sum_i d^{[i]}  \\ 
& \quad + \frac{1}{2} \bs{\Sigma}^{-1} \left( \sum_i d^{[i]}(\bs{z}^{[i]} -\mathbf{W}\bm{\phi}(\mathbf{s}))(\bs{z}^{[i]} -\mathbf{W}\bm{\phi}(\mathbf{s}))^{T} \right) \bs{\Sigma}^{-1} \\
& \quad = \mathbf{0}. 
\end{align*}

By solving for $\bm{\Sigma}$, we obtain
\begin{equation*}
\bm{\Sigma} =\frac{\sum_{i=1}^{N} d^{[i]}\left(\bm{u}^{[i]}-\bm{W}^{T} \bm{\phi}\left(\bm{s}^{[i]}\right)\right)\left(\bm{u}^{[i]}-\bm{W}^{T} \bm{\phi}\left(\bm{s}^{[i]}\right)\right)^{T}}{ \sum_i d^{[i]} }
\end{equation*}
Similar to Algorithm 1, we use
\begin{equation*}
Y = \frac{\left(\sum_{i=1}^{N} d^{[i]} \right)^2 - \sum_{i=1}^N (d^{[i]})^2 }{\sum_{i=1}^N d^{[i]}} 
\end{equation*}
as the demonimator to obtain an unbiased estimate of the covariance.

\end{document}